\documentclass[lettersize,journal]{IEEEtran}
\usepackage{amsmath,amsfonts}
\usepackage{algorithm}
\usepackage{algorithmicx}
\usepackage{algpseudocode}
\algdef{SE}[DOWHILE]{Do}{doWhile}{\algorithmicdo}[1]{\algorithmicwhile\ #1}%
\usepackage{array}
\usepackage{textcomp}
\usepackage{stfloats}
\usepackage{url}
\usepackage{verbatim}
\usepackage[caption=false,font=scriptsize,labelfont={scriptsize,bf}]{subfig}
\usepackage{graphicx}
\usepackage{cite}
\usepackage{multirow}
\usepackage{color}
\usepackage{tikz}
\usepackage{enumitem}
\usepackage{hyperref}
\definecolor{DentiBotColor}{rgb}{0 0.4470 0.7410}
\newcommand{\DentiBotLine}{\raisebox{2pt}{\tikz{\draw[-,DentiBotColor,solid,line width = 1pt](0,0) -- (5mm,0);}}}
\definecolor{PatientColor}{rgb}{0.6 0.6 0.6}
\newcommand{\PatientLine}{\raisebox{2pt}{\tikz{\draw[-,PatientColor,dotted,line width = 2pt](0,0) -- (5mm,0);}}}
\definecolor{ErrColor}{rgb}{1 0 0}
\newcommand{\ErrLine}{\raisebox{2pt}{\tikz{\draw[-,ErrColor,solid,line width = 1pt](0,0) -- (5mm,0);}}}
\definecolor{InsDeptTwlvColor}{rgb}{0.69 0.09 0.05}
\newcommand{\InsDeptTwlvLine}{\raisebox{2pt}{\tikz{\draw[-,InsDeptTwlvColor,solid,line width = 1pt](0,0) -- (5mm,0);}}}
\definecolor{InsDeptNineColor}{rgb}{1 0.84 0}
\newcommand{\InsDeptNineLine}{\raisebox{2pt}{\tikz{\draw[-,InsDeptNineColor,solid,line width = 1pt](0,0) -- (5mm,0);}}}
\definecolor{InsDeptSixColor}{rgb}{1 0.5 0}
\newcommand{\InsDeptSixLine}{\raisebox{2pt}{\tikz{\draw[-,InsDeptSixColor,solid,line width = 1pt](0,0) -- (5mm,0);}}}
\definecolor{InsDeptThreeColor}{rgb}{0.01 0.66 0.62}
\newcommand{\InsDeptThreeLine}{\raisebox{2pt}{\tikz{\draw[-,InsDeptThreeColor,solid,line width = 1pt](0,0) -- (5mm,0);}}}
\definecolor{InsDeptZeroColor}{rgb}{0.24 0.57 0.25}
\newcommand{\InsDeptZeroLine}{\raisebox{2pt}{\tikz{\draw[-,InsDeptZeroColor,solid,line width = 1pt](0,0) -- (5mm,0);}}}

\begin{document}

\title{DentiBot: System Design and 6-DoF Hybrid Position/Force Control for Robot-Assisted Endodontic Treatment}

\author{Hao-Fang Cheng, Yi-Ching Ho, and Cheng-Wei Chen 
\thanks{This work was supported in part by the National Science and Technology Council in Taiwan (Young Scholar Fellowship NSTC 111-2636-E-002-028 and 112-2628-E-002-021-MY3). Hao-Fang Cheng and Cheng-Wei Chen are with the Department of Electrical Engineering, National Taiwan University, Taipei, Taiwan. Yi-Ching Ho is with Division of Endodontics and Periodontology, Department of Stomatology, Taipei Veterans General Hospital, Taipei, Taiwan and Department of Dentistry, National Yang Ming Chiao Tung University, Taipei, Taiwan. }
\thanks{Corresponding author: Cheng-Wei Chen {\tt\small cwchenee@ntu.edu.tw}}}



\maketitle

\begin{abstract}
Robotic technologies are becoming increasingly popular in dentistry due to the high level of precision required in delicate dental procedures. Most dental robots available today are designed for implant surgery, helping dentists to accurately place implants in the desired position and depth. In this paper, we introduce the DentiBot, the first robot specifically designed for dental endodontic treatment. The DentiBot is equipped with a force and torque sensor, as well as a string-based Patient Tracking Module, allowing for real-time monitoring of endodontic file contact and patient movement. We propose a 6-DoF hybrid position/force controller that enables autonomous adjustment of the surgical path and compensation for patient movement, while also providing protection against endodontic file fracture. In addition, a file flexibility model is incorporated to compensate for file bending. Pre-clinical evaluations performed on acrylic root canal models and resin teeth confirm the feasibility of the DentiBot in assisting endodontic treatment.
\end{abstract}



\section{Introduction}
\label{sec:Intro}

Surgical robots have undergone rapid development in various clinical specialties, particularly in microsurgery, as robotics and automation technologies offer greater precision and repeatability compared to human surgeons \cite{gerhardus2003robot,mattos2016microsurgery}. Notable examples include robotic systems designed to assist intraocular surgery, such as the Steady Hand Manipulator \cite{taylor1999steady}, Preceyes \cite{de2016release}, and IRISS \cite{wilson2018intraocular}. Through the reduction of hand tremors and the incorporation of intraoperative imaging, robotic manipulators have successfully facilitated the performance of delicate vitreoretinal procedures during human trials \cite{edwards2018first}. Despite the progress in microsurgical robotics, the availability of robots specifically designed for automating dental surgery remains limited \cite{ahmad2021dental,liu2023robotics}. In this study, our objective is to design a robotic system that assists in dental endodontic treatment, commonly known as root canal treatment (RCT).

In recent years, dental robots have gradually gained attention \cite{van2021robot}. The FDA-approved Yomi manipulator stands as the pioneering example in this field \cite{bolding2021accuracy}. Dental robots, such as the Yomi and other systems developed at Hangzhou Normal University \cite{sun2014automated}, University of Hong Kong \cite{li2019compact}, Shanghai Jiao Tong University \cite{feng2022image}, and UCLA \cite{wang2022automation}, have been primarily designed to automate dental implantation procedures, including drilling and implant placement. Additionally, some systems have been developed specifically for crown preparation \cite{wang2014preliminary, yuan2016automatic}. The unique challenge in designing dental robots lies in the confined workspace of dental procedures, setting them apart from other microsurgeries. Unlike the need to minimize incision size in eye surgery, dental procedures do not require a remote center of motion (RCM) mechanism. Consequently, in contrast to the utilization of parallel kinematics mechanisms \cite{feng2022image}, serial mechanisms have been widely adopted in dental robotics \cite{sun2014automated, bolding2021accuracy, li2019compact, wang2022automation,yan2022optics}. Only a small number of systems still rely on the use of the RCM mechanism \cite{kim2009study, iijima2020development}.

Dental implantation typically involves following a predetermined motion path that has been carefully planned based on a preoperative computed tomography (CT) model \cite{zheng2007computer}. However, the patient remains conscious during dental implantation, as well as during many other dental procedures, as they are not placed under general anesthesia. Thus, accurate tracking of head and jaw motion becomes critical. Although 6-DoF passive manipulators have been utilized for real-time pose measuring \cite{bolding2021accuracy, sin2023development}, their significant physical footprint renders them unsuitable for narrow clinic spaces. On the other hand, visual-based tracking is widely applied for robot-patient alignment \cite{zheng2007computer, feng2022image, daon2014system, yan2022optics}, but it is less precise and susceptible to occlusion issues. While the direct attachment of a miniature robot onto the teeth \cite{Dong2007WIPAS} remains a technical challenge, improvements are needed in patient motion tracking for robot-assisted dental procedures. 

In 2022, the DentiBot was first implemented to assist with RCT \cite{cheng2022force}. RCT is a common procedure performed to address infected dental pulp and root canal system caused by deep cavities or dental trauma \cite{hulsmann2005mechanical,gutmann2010problem}. The RCT procedure comprises several important steps, including access opening, cleaning and shaping, and obturation. Among these steps, the cleaning and shaping process presents significant challenges and potential risks \cite{gutmann2010problem, jafarzadeh2007ledge}. Dentists use endodontic files, which can be manual or engine-driven \cite{liang2022evolution}, with various diameters to progressively insert and shape the root canal while removing pathogenic contents. The process continues until the file tip reaches the apex and the root canal achieves the desired taper and enlargement. Although engine-driven files reduce the operation time, it is crucial to avoid exceeding the axial torque and radial force limit of them, as doing so can result in file fractures \cite{sattapan2000defects}. Due to the absence of visual feedback inside the root canal, dentists traditionally perform this step manually, relying on their experience and tactile sensation to navigate the file. In other words, autonomous cleaning and shaping is very challenging with only visual-based position feedback. However, with the introduction of the DentiBot, there is potential for automation and enhanced precision in this critical step of the RCT procedure.

Previous studies have demonstrated the successful application of force guidance in the dental implant procedure, effectively compensating for slight misalignments caused by kinematic errors or patient motion \cite{kim2009study, feng2022image,wang2022automation}. This concept aligns with the principle of peg-in-hole assembly using force-guided robotic manipulators \cite{qiao1993robotic,wang2019robotic}. Building upon the same concept, the DentiBot incorporates 6 DoF force/torque feedback to autonomously adjust its insertion path when entering the root canal, ensuring the preservation of the endodontic file's integrity. The effectiveness of force guidance in assisting the cleaning and shaping step of RCT has been experimentally demonstrated \cite{cheng2022force}. However, the current performance of the DentiBot in this regard is not entirely satisfactory, as it exhibits alignment errors of up to 3 mm. This can be attributed primarily to the limited consideration given to the flexibility of endodontic files and the shape of the root canal.

Unlike dental implants, endodontic files possess flexibility that enables them to adapt to the curved structure of a root canal. However, this flexibility needs to be properly accounted for when applying force guidance. Additionally, the shape of the root canal exhibits a conical pattern, characterized by a wider entrance that gradually tapers as it approaches the apex \cite{lim1985validity,toosi2014virtual}. The endodontic file, being narrower than the root canal entrance, can have a small angular motion inside the root canal without making contact with the canal wall. In addition, rapid movements by the patient can also lead to saturation of the force/torque sensor. To achieve autonomous RCT and ensure precise patient alignment, it is imperative to integrate both position and force control strategies. This integration will allow for better adaptation to the curved root canal and enable the DentiBot to perform RCT autonomously while accurately aligning with the patient.

\begin{figure}[!t]
  \centering
        \includegraphics[width=8.5cm]{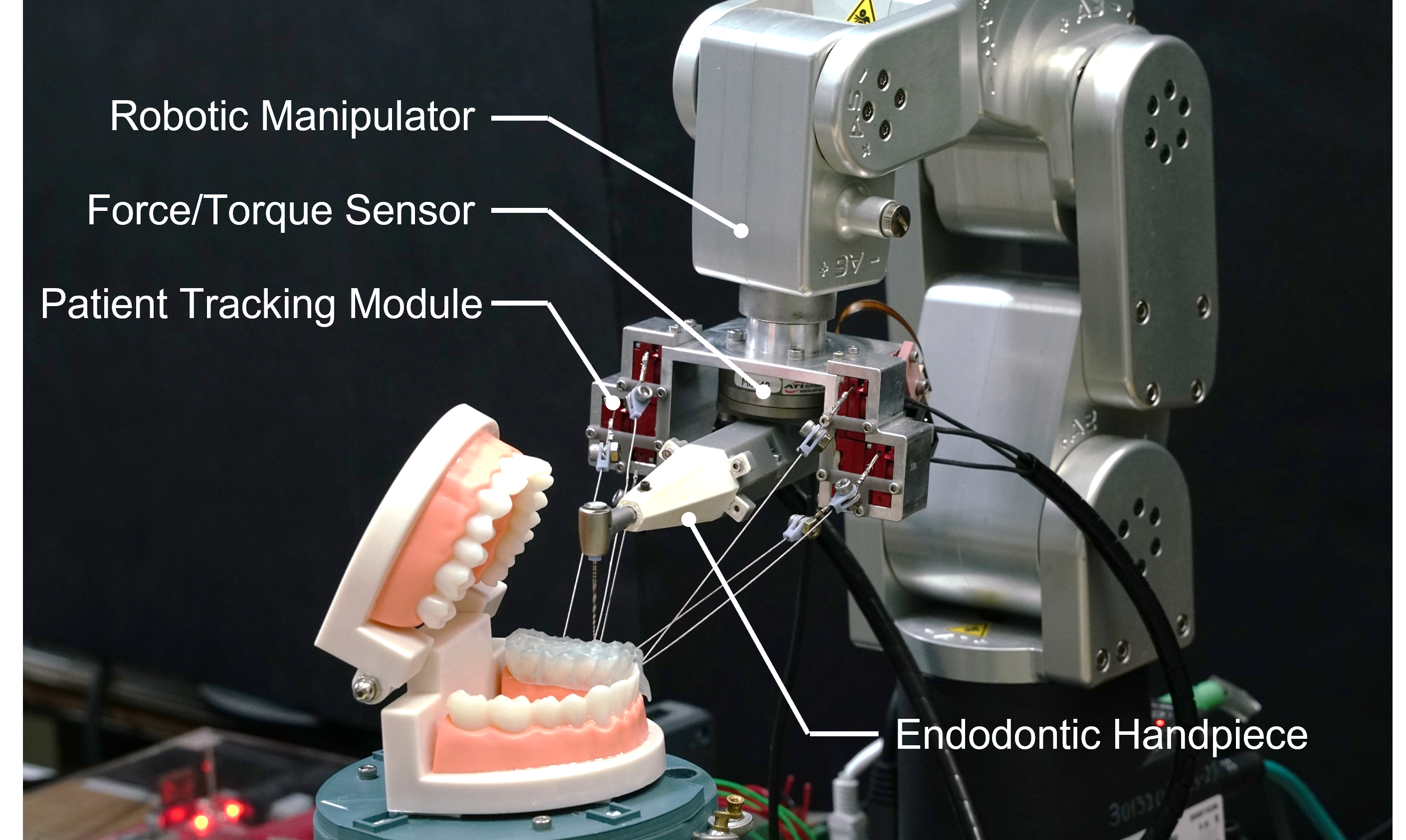}
        \vspace{-2mm}
        \caption{DentiBot, designed for performing robot-assisted endodontic treatment, consists of a 6-DoF robotic manipulator, a 6-axis force/torque sensor, an endodontic handpiece, and a string-based patient tracking module (PTM). [\href{https://www.youtube.com/watch?v=cxKhz5po8vI}{YouTube Video}] }
        \label{fig:dentibot}
        \vspace{-5mm}
\end{figure}

In this study, we have upgraded the DentiBot (Fig. \ref{fig:dentibot} [\href{https://www.youtube.com/watch?v=cxKhz5po8vI}{YouTube Video}]) by incorporating a string-based Patient Tracking Module (PTM). The PTM utilizes six string potentiometers to establish a connection between the robotic manipulator and the patient's jaw. By measuring the lengths of these strings, we can estimate the relative pose between the DentiBot and the surgical site. The design of the PTM, tailored for the DentiBot, offers accurate pose measurements in the designiated workspace. The risk of wire crossing, which is a common issue in string-based measuring methods, is analyzed and reduced though a Monte Carlo simulation. Besides, we have developed a model to compensate for endodontic file bending. This compensation enables a more accurate alignment and minimizes the force applied to the file. Furthermore, we have implemented a hybrid 6-DoF position/force control strategy. By combining position and force feedback control, the DentiBot can overcome the limitations associated with relying solely on each type of feedback. This integrated approach enables the DentiBot to adapt better to the flexibility of endodontic files and effectively manage the curved root canal during autonomous cleaning and shaping. Most importantly, this advancement in control methodology ensures precise alignment between the robot and the patient, paving the way for autonomous RCT procedures with enhanced accuracy and safety.

Evolved from our previous work \cite{cheng2022force}, in which only force feedback is applied to assist endodontic treatment, the contributions of this article are summarized as follows:
\vspace{-1mm}
\begin{enumerate}
\item 	Identification of clinical requirements: We carefully analyze the clinical needs for endodontic treatment and define specific specifications for the robotic surgical system. This ensures a tailored solution for robot-assisted endodontic procedures.
\item	String-based PTM for jaw motion tracking: We create a string-based PTM that seamlessly integrates with the DentiBot to track the patient's jaw motion accurately. This compact design minimizes interference and provides essential feedback for precise robotic control.
\item	Compensation for endodontic file bending: To address challenges related to file bending during RCT, we develop a compensation mechanism. It reduces the force applied to the file, mitagating the risk of file fracture, and enhancing treatment safety.
\item	Hybrid 6-DoF position/force control: This hybrid control strategy enables the alignment for invisible root canals and maintenance of the robot's pose relative to the patient, thereby guaranteeing the reliable performance of DentiBot.
\end{enumerate}
\vspace{-1mm}

The rest of this article is organized as follows: the analysis of clinical requirements and system specifications are provided in Section \ref{sec:Spec}; the system design of the DentiBot, including the kinematics and integration of the 6-DoF robotic manipulator and the force/torque sensor, is described in Section \ref{sec:SystDesign}; The design and performance analysis of the PTM are provided in Section \ref{sec:PTM}; Section \ref{sec:Intra-guid} elaborates the hybrid position/force control strategy implemented on the DentiBot; the visualization and user interface design are given in Section \ref{sec:Visualize}; Section \ref{sec:Exp} demonstrates the evaluation of the prototypical system; the concluding remarks are given in Section \ref{sec:conclu}.


\section{Clinical Requirements and Specifications}
\label{sec:Spec}
The process of endodontic treatment is briefly introduced in this section, followed by necessary requirements and specifications for a robotic system designed to execute the treatment procedures autonomously.

\subsection{Procedure of Endodontic Treatment}

RCT or endodontic treatment is a dental procedure performed to save a severely damaged or infected tooth \cite{cohen1998pathways}. Teeth consist of layers including enamel, dentin, and a pulp chamber containing dental pulp with blood vessels, nerves, and connective tissues. When the dental pulp becomes infected or damaged due to factors like decay, cracks, or trauma, it can lead to pain and abscess formation. RCT aims to remove the infected pulp, clean and shape the root canals, and fill them to prevent reinfection. The expected outcome is pain relief, elimination of infection, and preservation of the natural tooth, allowing it to function normally with proper care and potentially lasting a lifetime.

\begin{figure}
  \centering
        \includegraphics[width=8.5cm]{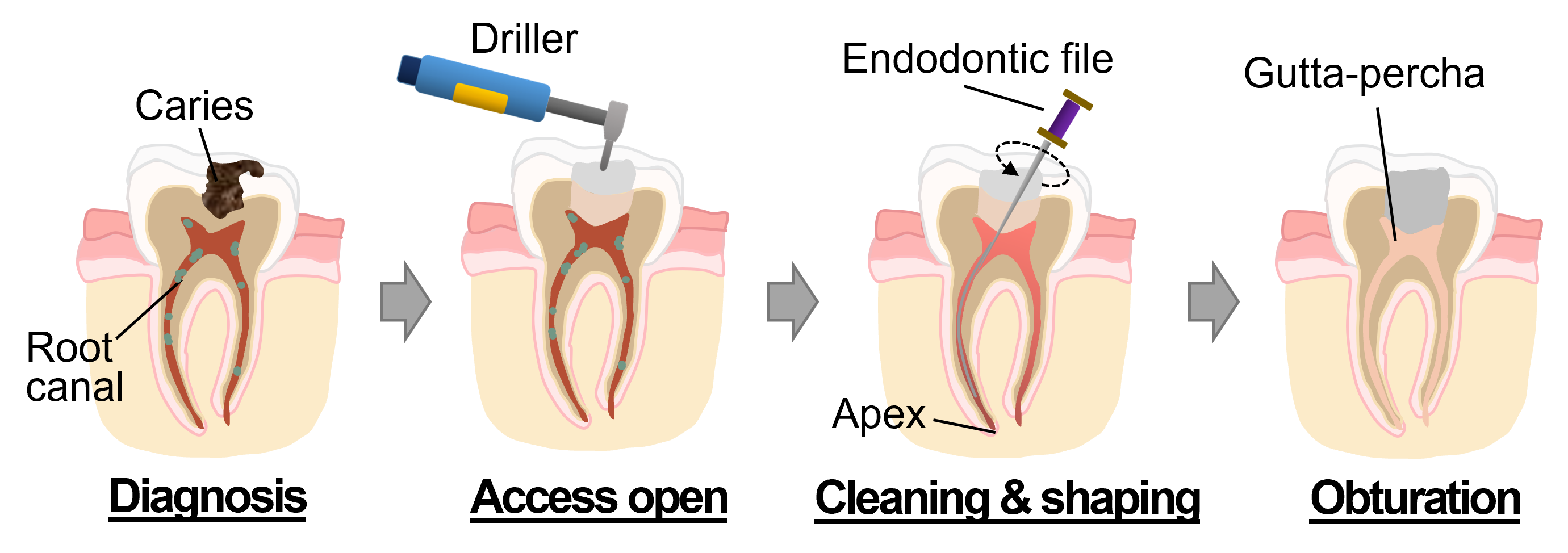}
        \vspace{-4mm}
        \caption{Procedure of the endodontic treatment. In the case of root canal infection, the dentist will remove the affected tooth and create an access open. Subsequently, endodontic files of increasing diameters are used to clean and shape the root canal. Once the cleaning and shaping step is complete, Gutta-percha will be used to obturate the root canal.}
        \label{fig: endodontic treatment}
\vspace{5mm}
  \centering
        \includegraphics[width=8cm]{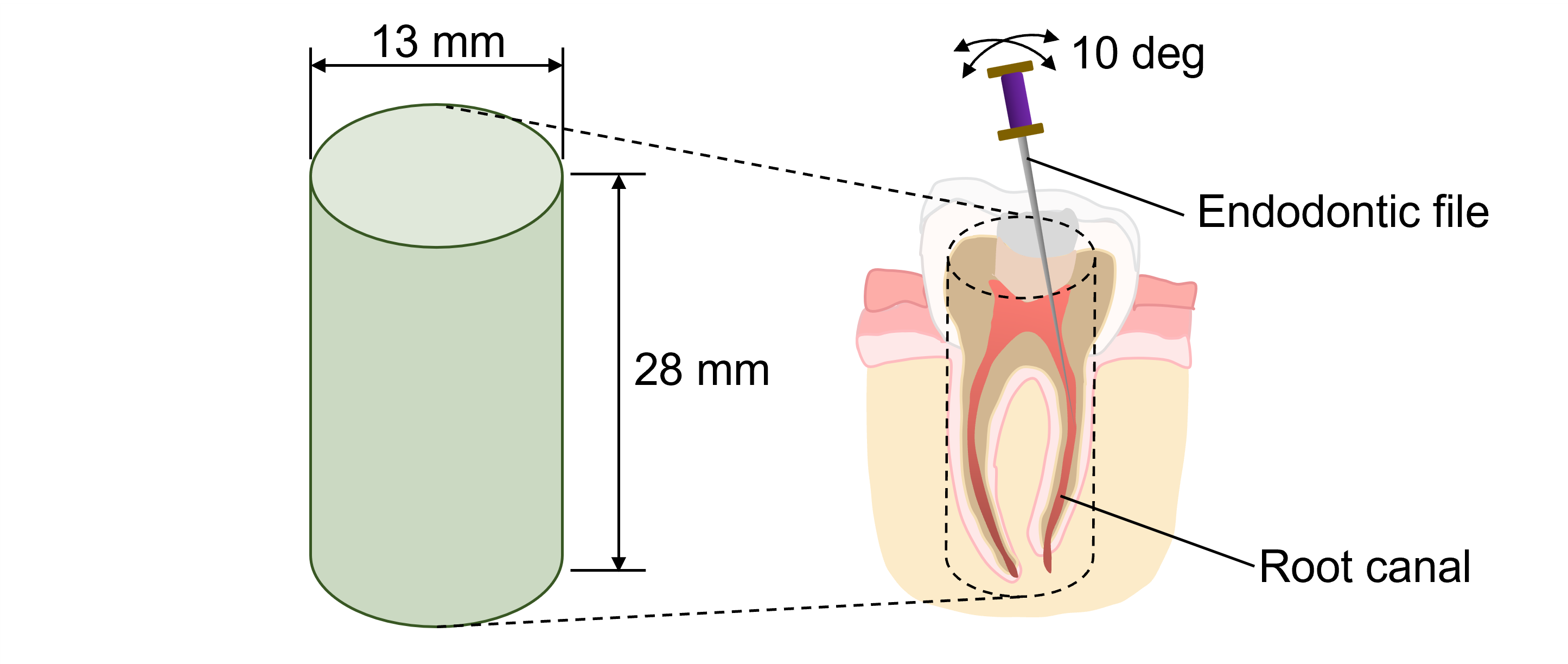}
        \vspace{-4mm}
        \caption{RCT needs a cylindrical workspace, with a radius of $6.5$ mm and a length of $28$ mm, for cleaning and shaping the root canal cavity. To facilitate endodontic file insertion and manipulation during the treatment, file reorentation in roll and pitch angles should be allowed up to 10 degrees.}
        \label{fig:tooth_workspace_defi}
        \vspace{-3mm}
\end{figure}

The procedure of endodontic treatment involves several key surgical steps, as illustrated in Fig. \ref{fig: endodontic treatment}. Following the diagnosis of the need for RCT, local anesthesia is administered to provide numbing of the tooth and its surrounding area, ensuring a painless experience for the patient. Subsequently, a dental bur is employed to create an access opening in the tooth, typically through the crown \cite{adams2014access}. This access point allows the dentist to reach the pulp chamber and root canals for further treatment.

After completing the initial steps, the cleaning and shaping stage, crucial in endodontic treatment, ensues to remove the damaged pulp tissue and infected dentin from the root canal systems. Dentists employ specialized endodontic files during this step, often utilizing rotary files with sharp cutting edges. By employing a careful in-and-out motion, the endodontic files are manipulated until they reach the apex of the root canal. To determine the precise location of the apex, an apex locator or radiographs may be used. Given the limited visibility, dentists heavily rely on their tactile sense to ensure accurate adjustment of the file insertion path. Gradually, the root canal is enlarged by using files of increasing sizes, which are determined by the dentist's judgment based on the unique characteristics of each root canal. 

Instrument fracture is a significant concern during endodontic treatment \cite{gutmann2010problem}. Removing a separate file in the root canal is technically sensitive. There is another failure circumstance called ledge, which means that the file drills in the wrong path and turn out to be stuck in the root canal. File misalignment and ledging can result in undesired remains of infected tissue in the root canal system and overpreparation of the root dentin. Hence, it is essential to align the file with the root canal and reduce the probability of instrument fracture and ledging. 

After cleaning and shaping step, the dentist fills the root canals with biocompatible materials, usually gutta-percha and sealers, to ensure complete sealing of the root canals. Adequate coronal restoration is also necessary after obturation. Obturation and coronal restoration prevent microorganisms' recontamination and byproducts from entering the root canal system. Finally, the dentist will fabricate a crown or other restoration to protect the tooth.

Overall, endodontic treatment, particularly the cleaning and shaping step, is a complex and delicate procedure that requires precise execution to ensure optimal outcomes for patients. However, the patients would be only locally anesthetized around their oral tissues. They still can arbitrarily move their head and jaw. Conventionally, the dentist would hold the patient's jaw and compensate for any movement through their well-trained experience.

In this study, we focus on the automation of the cleaning and shaping step in RCT, helping the dentist with real-time force sensing and delicate movements by the robot. The following clinical requirements have been identified:
\begin{enumerate}[label=(R\arabic*)]
\item Achieving six degree-of-freedom (DoF) motions of the flexible endodontic file within the pulp chamber and root canals.
\item Implementing force guidance to address the absence of visual feedback and the risk of file fracture inside the curved root canal.
\item Tracking patient's jaw motion during the procedure to maintain a fixed relative pose between the robot and the patient.
\end{enumerate}

\subsection{System Specifications}
In order to perform robot-assisted endodontic treatment, several system specifications must be determined. Firstly, the required workspace is defined as a cylinder (see Fig. \ref{fig:tooth_workspace_defi}) to accommodate the size and shape of teeth. Considering the average diameter of molar teeth, approximately $10.7$ mm \cite{toothsize}, a cylinder diameter of $13$ mm is chosen to account for outlier cases. The cylinder height is determined based on the length of root canals. Canine teeth have the longest root canal, approximately $24$ mm \cite{rootcanal}. Therefore, the cylinder height is set to $28$ mm to provide sufficient surgical depth. 

Considering the range of attitude in which the robot operates is also crucial. Before inserting the endodontic file, the robotic manipulator can be manually aligned with the target root canal. Therefore, the critical range of angles pertains to the movement of the file within a single root canal. Research indicates that there is a deviation of $3.8-5.0$ degrees in the attitude of the endodontic file during RCT \cite{zehnder2016guided}. This deviation is primarily caused by the conical shape of root canals, which have an entrance width of up to $1.5$ mm and gradually narrow down to $0.2-0.6$ mm at the apex \cite{lee2020three}. To maintain precision and minimize the risk of file fracture, it is advisable to allow for the manipulation angle at least $10$ degrees for the roll and pitch axis. On the other hand, the yaw axis, or file rotation, shall have a full range of $360$ degrees.

In this study, NiTi (Nickel-Titanium) rotary files (ProTaper Universal SX--F3, Dentsply Sirona, Ballaigues, Switzerland), with the length of $21$ mm and the apical diameter of $0.18-0.30$ mm, are used in autonomous cleaning and shaping step. NiTi files are valued for their flexibility \cite{haapasalo2013evolution}. In general, NiTi files have a maximum bending angle more than $45$ degrees \cite{kwak2021effects}. They can navigate curved root canal pathways with ease, reducing the risk of ledging \cite{pettiette2001evaluation}. ProTaper files are compatible with motorized handpieces or automated systems. The rotation speed is recommended at $150-350$ rpm. In addition, the mean apical force and axial torque need to be limited to $3.9$ N and $12$ mN$\cdot$m, respectively, to ensure the safety of the treatment \cite{martins2020mechanical}.

Patient movement is a critical consideration in robot-assisted RCT. On average, the patient's linear displacement speed during the procedure is approximately $1$ mm/s, with a maximum speed of $2.5$ mm/s \cite{wang2022automation}. The angular displacement speed is around $0.5$ deg/s, with a maximum speed of $1$ deg/s. Precise tracking of the patient's motion within these velocity ranges is crucial to ensure safety. In this study, we aim to reduce the tracking error of the DentiBot to $2$ mm, a significant improvement from its previous level of up to $5$ mm \cite{cheng2022force}. In the event that the patient exceeds the tolerable movement speed, the robot shall be programmed to immediately halt the surgical procedure, allowing the dentist to regain full control and ensure the patient's safety.

In summary, the following specifications have been defined for the DentiBot:
\begin{enumerate}[label=(S\arabic*)]
\item Workspace: a cylinder working volume with the height of $28$ mm and diameter of $13$ mm. The minimum allowable roll and pitch angles are $10$ degrees.
\item Maximum apical force and axial torque exerted to the endodontic file: $3.9$ N and $12$ mN$\cdot$m. 
\item Error tolerance for patient motion tracking: $2$ mm with the maximum velocity of $2.5$ mm/s.
\end{enumerate}


\section{Robotic System Design}
\label{sec:SystDesign}
In this section, we elaborate on the system design of the DentiBot, which is aimed at meeting the clinical requirements and specifications of performing RCT.


\subsection{System Overview}
\begin{figure}[!b]
\vspace{-4mm}
  \centering
        \includegraphics[width=8cm]{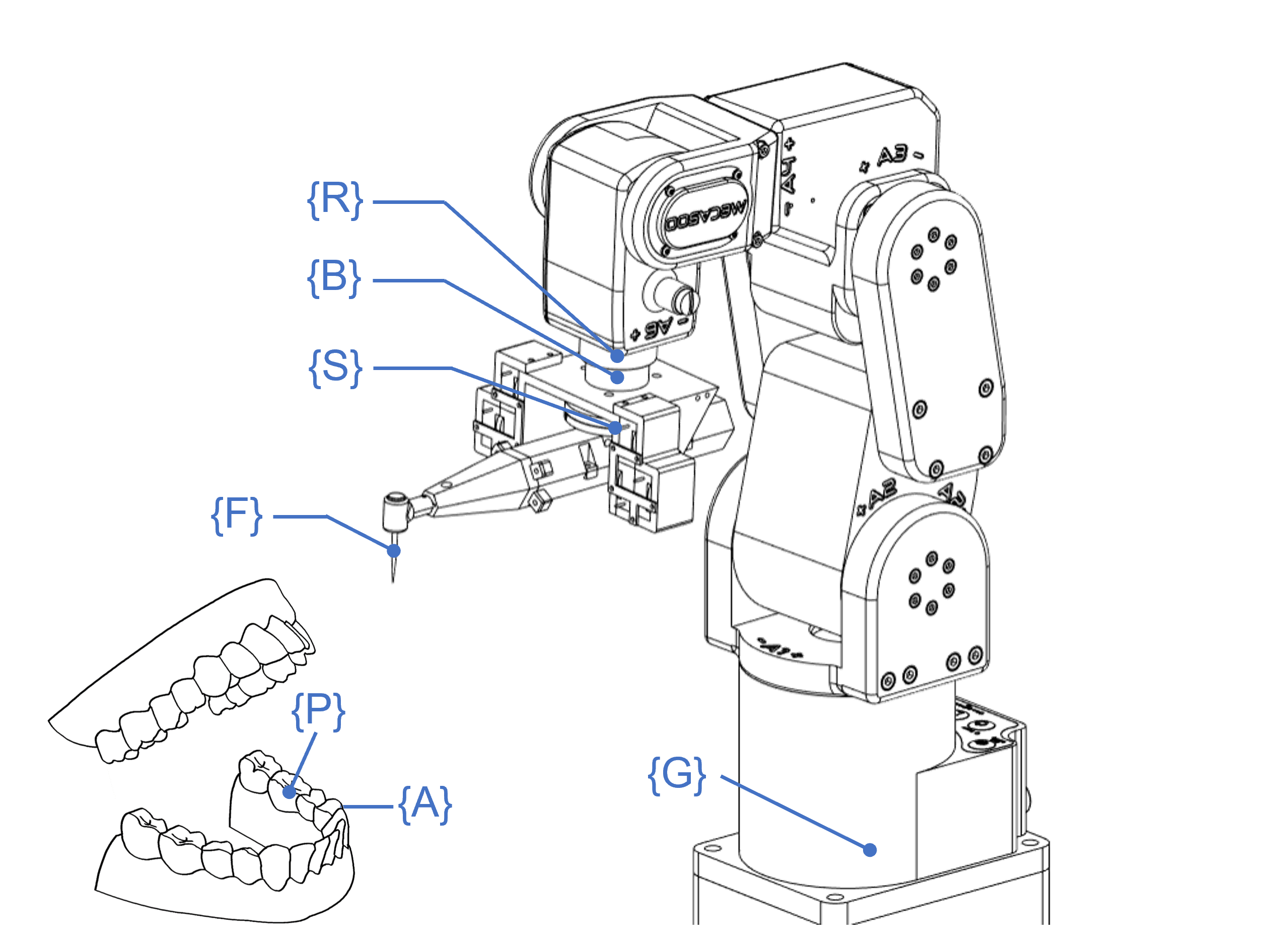}
        \caption{Coordinate systems of the Dentibot, where $\{G\}$ and $\{R\}$ represent the ground and sixth link of the 6-DoF robotic manipulator, respectively; $\{A\}$ and $\{B\}$ denote the anchor and base of the PTM, respectively; $\{S\}$ denotes the measurement coordinate of the force/torque sensor; $\{F\}$ is locating at the endodontic file; and $\{P\}$ is at the location of the infected tooth.}
        \label{fig:dentibot_frame}
\end{figure}

The components of the DentiBot include a 6-DoF robotic manipulator, an endodontic handpiece, a 6-axis force/torque sensor, and a string-based PTM (as shown in Fig. \ref{fig:dentibot}). Each component has its own coordinate system, as defined in Fig. \ref{fig:dentibot_frame}. The robotic manipulator performs the surgical operations like a dentist. The ground link of the manipulator, represented as $\{G\}$, serves as a fixed frame for the entire kinematic chain. The sixth link of the manipulator is denoted as $\{R\}$ and is fixedly connected to the base frame of the PTM, $\{B\}$. Below the base of the PTM, the force/torque sensor is attached to measure the contact between the endodontic file (denoted as $\{F\}$, installed on the dental handpiece) and the root canal (denoted as $\{P\}$). The force and torque measurements are output in the sensor's frame $\{S\}$. While the dental handpiece drives the rotational motion of the endodontic file, the robotic manipulator maneuvers the file orientation and actuates the in-and-out motion to perform the cleaning and shaping of root canal systems.

On the patient side of the PTM, a teeth brace is mounted on the patient's jaw and represented as $\{A\}$. The PTM estimates the coordinate transformation between $\{A\}$ and $\{B\}$ in real time. In practice, the transformation between $\{R\}$ and $\{B\}$ is already known in the mechanical design. The transformation between $\{A\}$ and $\{P\}$ can be acquired through 3D medical imaging. As a result, the relative pose between the robot and the target tooth of the patient, i.e., the coordinate transformation between $\{R\}$ and $\{P\}$, is intraoperatively obtained for patient tracking. In the following subsections, we will provide an explanation of the kinematics and calibration procedures for both the robotic manipulator and the force/torque sensor. The detailed design and analysis of the string-based PTM are given in Section \ref{sec:PTM}.

\subsection{6-DoF Robotic Manipulator and Dental Handpiece}
A 6-DoF robotic manipulator (Meca500, MecaDemic, Montreal, Canada) carrying a motorized dental handpiece is employed to fulfill both the clinical requirement (R1) and the workspace specification (S1). Equipped with a motorized dental handpiece, the robotic manipulator positions the endodontic file within the designated cylindrical workspace with an accuracy of $0.1$ mm. The endodontic file can be easily replaced with various lengths and diameters to accommodate different treatment demands. 

To establish the coordinate transformation between the robot frame $\{R\}$ and endodontic file frame $\{F\}$, the tool center point (TCP) calibration \cite{hallenberg2007robot,yang2017four} is applied. Note that $\{F\}$ is placed at the top of the screw-shape working part of the file, rather than the file tip. Thus, two endodontic files with different lengths are required in this calibration process.

The longer endodontic file is first installed on the dental handpiece. By manually commanding the robotic manipulator, the file tip is positioned to a fixed point with four different orientations. The relative poses between $\{G\}$ and $\{R\}$ at different orientations are obtained using the Denavit-Hartenberg (D-H) Method \cite{fu1983robotics} and recorded. Next, the translational vector between $\{R\}$ and the file tip, denoted as $^\mathrm{R}\mathbf{t}_\mathrm{F1}\in \mathbb{R}^3$, is solved in the following least squares problem:
\begin{equation}
     \begin{bmatrix}
 _\mathrm{R}^\mathrm{G}\mathbf{R}(1)-{_\mathrm{R}^\mathrm{G}\mathbf{R}(2)} \\
 _\mathrm{R}^\mathrm{G}\mathbf{R}(1)-{_\mathrm{R}^\mathrm{G}\mathbf{R}(3)}\\
_\mathrm{R}^\mathrm{G}\mathbf{R}(1)-{_\mathrm{R}^\mathrm{G}\mathbf{R}(4)}
\end{bmatrix}
{^\mathrm{R}\mathbf{t}_\mathrm{F1}} =
 \begin{bmatrix}
 ^\mathrm{G}\mathbf{t}_\mathrm{R}(2)- {^\mathrm{G}\mathbf{t}_\mathrm{R}(1)} \\
 ^\mathrm{G}\mathbf{t}_\mathrm{R}(3)- {^\mathrm{G}\mathbf{t}_\mathrm{R}(1)}\\
 ^\mathrm{G}\mathbf{t}_\mathrm{R}(4)- {^\mathrm{G}\mathbf{t}_\mathrm{R}(1)}
\end{bmatrix}
,
\end{equation}
where $_\mathrm{R}^\mathrm{G}\mathbf{R}(j)$ and $^\mathrm{G}\mathbf{t}_\mathrm{R}(j)$, $j \in \{1,2,3,4\}$, represent the rotational matrix and translational vector, respectively, obtained from four different orientations in the TCP calibration process.

By repeating the same TCP calibration procedure on another shorter endodontic file, $^\mathrm{R}\mathbf{t}_\mathrm{F2}$ is obtained. Once completed, the rotational matrix $^\mathrm{R}_\mathrm{F}\mathbf{R}$ is calculated based on the directional vector ${^\mathrm{R}\mathbf{t}_\mathrm{F1}}-{^\mathrm{R}\mathbf{t}_\mathrm{F2}}$. The translational vector $^\mathrm{R}\mathbf{t}_\mathrm{F} $ is then derived by
\begin{equation}
  ^\mathrm{R}\mathbf{t}_\mathrm{F} = 
  \ ^\mathrm{R}\mathbf{t}_\mathrm{F1} - (\left \| ^\mathrm{F}\mathbf{t}_\mathrm{F1} \right \|_2 )\frac{(^\mathrm{R}\mathbf{t}_\mathrm{F1}-{^\mathrm{R}\mathbf{t}_\mathrm{F2}})}{\left \| (^\mathrm{R}\mathbf{t}_\mathrm{F1}-{^\mathrm{R}\mathbf{t}_\mathrm{F2}}) \right \|_2},
\end{equation}
where $\left \| ^\mathrm{F}\mathbf{t}_\mathrm{F1} \right \|_2$ denotes the length of the first file's working part.

The robotic manipulator accepts joint velocity commands $\dot{\mathbf{q}}_\mathrm{cmd} \in \mathbb{R}^6$. The geometry Jacobian matrix $\mathbf{J}$ is applied to convert $^\mathrm{F}\mathbf{\dot{p}}_\mathrm{cmd} \in \mathbb{R}^6$, the velocity command along the linear ($x$,$y$,$z$) and rotary ($\phi$,$\psi$,$\theta$) axes with respect to the endodontic file frame $\{F\}$, to $\dot{\mathbf{q}}_\mathrm{cmd}$, the joint velocity command \cite{lewis1993control}:
\begin{equation}
\label{eq:jacobian}
    \begin{split}
     \dot{\mathbf{q}}_\mathrm{cmd}
     = {\mathbf{J}^{-1}}  \begin{bmatrix}
 ^\mathrm{R}_\mathrm{F}\mathbf{R} & \boldsymbol{0}\\ 
 \boldsymbol{0} & ^\mathrm{R}_\mathrm{F}\mathbf{R}
 \end{bmatrix}
 {^\mathrm{F}\mathbf{\dot{p}}_\mathrm{cmd}}
    \end{split},    
\end{equation}   

\begin{equation}
\mathbf{J} =
\begin{bmatrix}
\frac{\partial{^\mathrm{G}\mathbf{p}_\mathrm{R}}}{\partial{q_1}}\
&\frac{\partial{^\mathrm{G}\mathbf{p}_\mathrm{R}}}{\partial{q_2}}\
&\cdots\
&\frac{\partial{^\mathrm{G}\mathbf{p}_\mathrm{R}}}{\partial{q_6}}
\end{bmatrix}. 
\label{eq:Jacob}
\end{equation}
Note that $q_i \in \mathbb{R}$, $i = \{1, 2, ..., 6\}$, represents the angle of the $i$-th joint. The pose vector from the ground link to the six link, $^\mathrm{G}\mathbf{p}_\mathrm{R} \in \mathbb{R}^6$, is obtained from the forward kinematics of the 6-DoF robotic manipulator.


\subsection{Force/Torque Sensor Calibration}
In order to facilitate intraoperative force guidance and fulfill the clinical requirement (R2) and force/torque limit specification (S2), we employ a 6-axis force/torque sensor (Mini40, ATI Industrial Automation, Apex, NC) that connects the sixth link of the robotic manipulator and the motorized dental handpiece to measure the force and torque applied to the endodontic file. Both force and torque measurements are originally obtained in the sensor's frame $\{S\}$ with the resolution of $0.01$ N and $0.25$ mN$\cdot$m, respectively.

Before applying the force/torque measurements for force guidance, it is essential to conduct gravity compensation. As the orientation of the force/torque sensor and dental handpiece relative to the ground undergoes changes during the operation, compensation for the weight of the dental handpiece becomes necessary. We employ the compensation method introduced by Vougioukas \textit{et al.} \cite{vougioukas2001bias}:
\begin{align}
\label{eq: gravitycomp_force}
\begin{split}
^\mathrm{S}\boldsymbol{f}_s
&= \
^\mathrm{S}\boldsymbol{f}_0
- \
^\mathrm{S}_{R}\mathbf{R}
^\mathrm{R}_{G}\mathbf{R} \,
\boldsymbol{w}_{h},
\end{split}\\
\begin{split}
^\mathrm{S}\boldsymbol{\tau}_s
&= \
^\mathrm{S}\boldsymbol{\tau}_0
- \
\boldsymbol{r}_{h} \,\times\,
(^\mathrm{S}\boldsymbol{f}_0
-{^\mathrm{S}\boldsymbol{f}_s}),
\end{split}
\label{eq: gravitycomp_torque}
\end{align} 
where $^\mathrm{S}\boldsymbol{f}_s$ and $^\mathrm{S}\boldsymbol{f}_0$ denote the compensated and raw force measurements, respectively, in the force sensor's coordinate frame $\{S\}$. $^\mathrm{S}\boldsymbol{\tau}_s$ and $^\mathrm{S}\boldsymbol{\tau}_0$ represent the compensated and raw torque measurements, respectively. $\boldsymbol{w}_{h}$ and $\boldsymbol{r}_{h}$ are the unknown weight and centroid position of the dental handpiece, respectively. $^\mathrm{R}_{G}\mathbf{R}$ can be obtained from forward kinematics of the robotic manipulator. $^\mathrm{S}_{R}\mathbf{R}$, however, includes an unknown mounting error along $\theta$-axis of the sensor. 

To determine the unknown parameters like $\boldsymbol{w}_{h}$ and $^\mathrm{S}_{R}\mathbf{R}$, a least squares method is applied. First, we assign certain poses to the robotic manipulator and record both $^\mathrm{R}_{G}\mathbf{R}$ and $^\mathrm{S}\boldsymbol{f}_0$ at each pose. By assuming there is no applied force, i.e., $^\mathrm{S}\boldsymbol{f}_s$ equals zero, and reorganizing the right hand side of Eq. \ref{eq: gravitycomp_force}, the unknown variables are obtained. Similarly, $\boldsymbol{r}_{h}$ is identified from another least squares problem using Eq. \ref{eq: gravitycomp_torque}.

Furthermore, as the sensor frame $\{S\}$ does not align with the endodontic file frame $\{F\}$, another coordinate transformation from $\{S\}$ to $\{F\}$ is required:
\begin{align}
    ^\mathrm{F}\boldsymbol{f}_s &=\, ^\mathrm{F}_\mathrm{R}\mathbf{R}\ ^\mathrm{R}_\mathrm{S}\mathbf{R}\ ^\mathrm{S}\boldsymbol{f}_s,\\
    \begin{split}
    ^\mathrm{F}\boldsymbol{\tau}_s 
    &=\,
    ^\mathrm{F}_\mathrm{S}\mathbf{R} ^\mathrm{S}\boldsymbol{\tau}_s
    \,+\,      {^\mathrm{F}\!\mathbf{t}_\mathrm{S}}\,\times\,^\mathrm{F}\boldsymbol{f}_s,
    \end{split}
\end{align}
where $^\mathrm{F}_\mathrm{R}\mathbf{R}$ and $^\mathrm{R}_\mathrm{S}\mathbf{R}$ are obtained from the TCP calibration and gravity compensation processes, respectively. Note that the torque sensed in the file frame $\{F\}$ comprises both the torque sensed in frame $\{S\}$ and an extra torque resulting from the force ${^\mathrm{F}\boldsymbol{f}s}$ applied to the lever ${^\mathrm{F}\!\mathbf{t}_\mathrm{S}}$, which is determined from the CAD model.


\section{String-Based Patient Tracking Module}
\label{sec:PTM}
The DentiBot utilizes a string-based PTM to estimate and regulate the relative pose between the patient and the robot during the surgical procedure. This functionality enables the fulfillment of both the clinical requirement (R3) and the patient tracking specification (S3). The design and kinematic analysis of the PTM are given in this section.


\subsection{Module Design}
To minimize the size of the passive pose measuring device required, we employ a string-based parallel kinematic mechanism. In essence, this mechanism calculates the relative pose between two rigid bodies by measuring the lengths of six strings connecting them. Previous research has introduced different configurations, specifically the 3-2-1 type and the 2-2-2 type. In the 3-2-1 type, three ball joints are affixed to the mobile object, with three, two, and one string(s) connected to each of these three joints, respectively \cite{geng19943,thomas2005performance}. The relative pose can be determined analytically using trilateration, which has been applied in studying human walking patterns \cite{ottaviano2010application}. 

To address the issue of multiple solutions encountered in the 3-2-1 type measuring device, an alternative 2-2-2 type has been suggested \cite{jeong1999kinematics}. In this configuration, the relative pose is determined through numerical methods like the Newton-Raphson method. However, when the device's footprint is minimized, it also reduces the variation of the six string lengths with regard to the change in relative pose. Consequently, the estimation accuracy is sensitive to measurement noises.

\begin{figure}[!t]
\centering
\subfloat[Design of the PTM]{%
  \includegraphics[width=8cm]{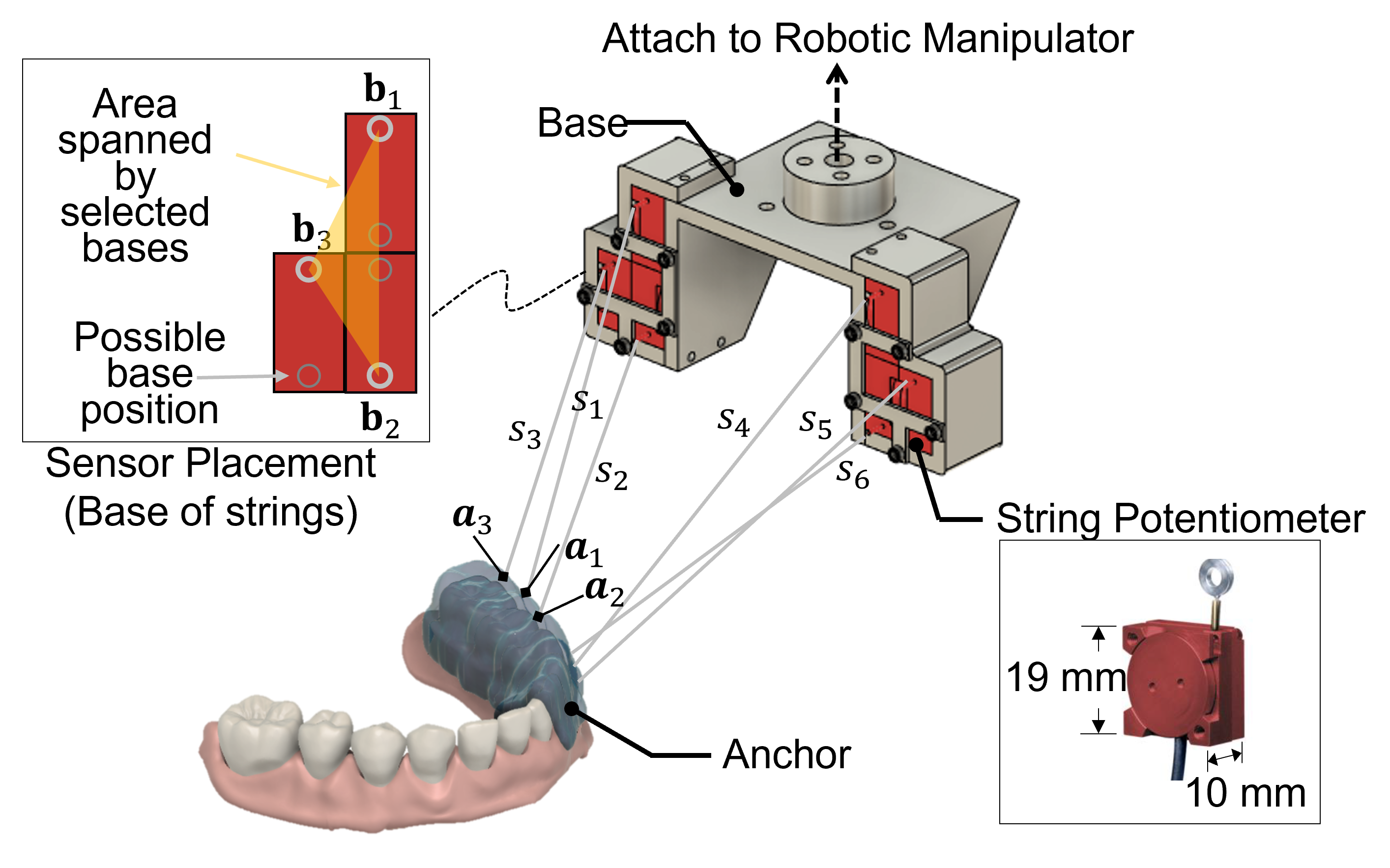}%
  \label{fig:PTM_Base}%
}\quad
\vspace{3mm}
\subfloat[Teeth brace and string anchors]{%
  \includegraphics[width=8cm]{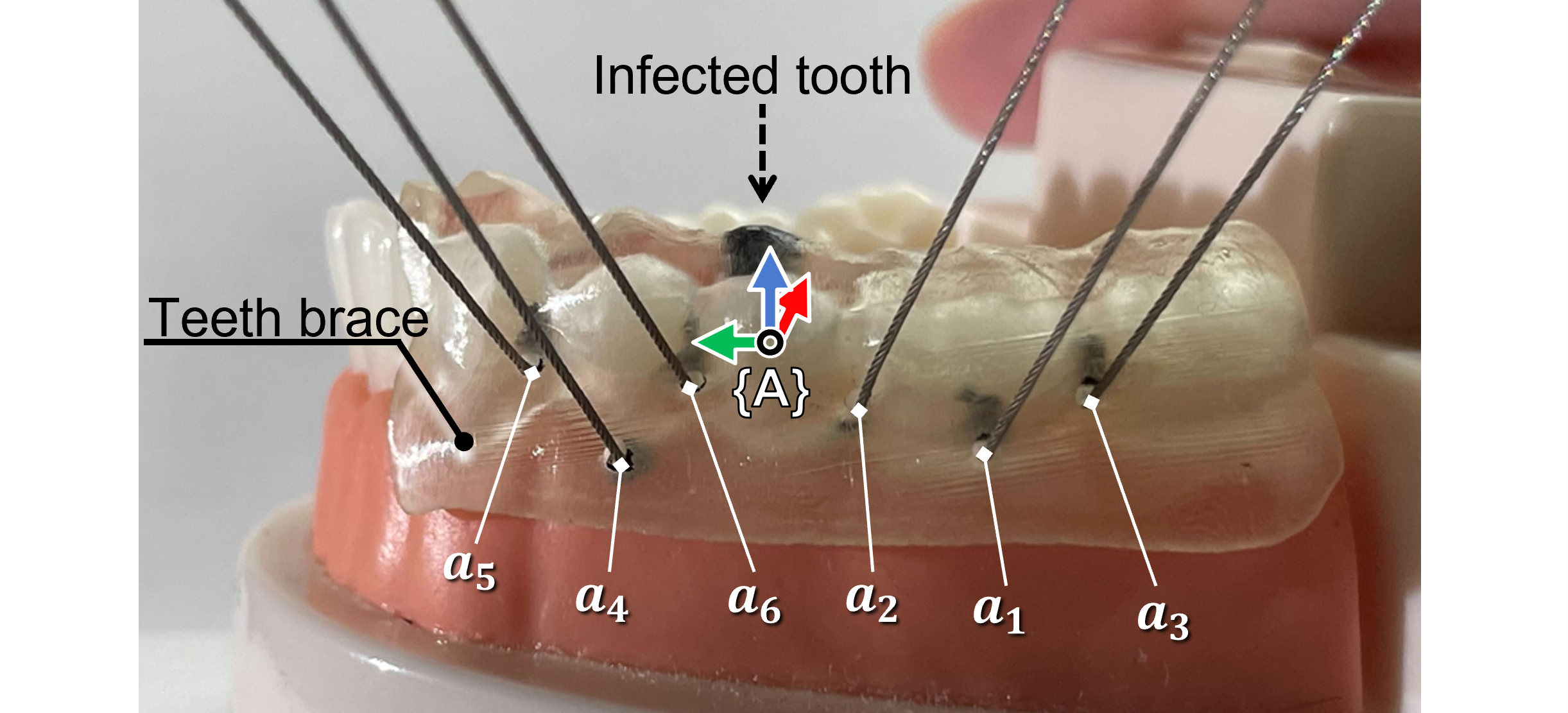}%
  \label{fig:PTM_Anchor}%
}
\caption{Design of the string-based patient tracking module (PTM), which includes the string potentiometers on the base adapter and the anchors on the teeth brace. (a) With six strings ($\mathfrak{s}_i, i \in \{1,2,\cdots,6\}$) connecting the DentiBot and the patient, the PTM estimates the relative position and orientation based on the string lengths $l_i, i \in \{1,2,\cdots,6\}$, which are measured by the potentiometers. (b) The anchor positions of the string, $\mathbf{a}_i, i \in \{1,2,\cdots,6\}$, are fixed on the teeth brace. Note that the anchor positions are selected to reduce the risk of string-crossing. The base positions of the string, $\mathbf{b}_i, i \in \{1,2,\cdots,6\}$, are chosen to minimize the footprint of the PTM while maximize the area spanned by the base points.} 
\label{fig:ptm_module}
\vspace{-5mm}
\end{figure} 

In this study, we employ the ``1-1-1-1-1-1'' type, where each string connects to a separate ball joint, to develop a passive pose measuring device aimed at enhancing estimation accuracy. The relative pose is still determined using the Newton-Raphson method, which theoretically offers up to 40 potential solutions \cite{raghavan1993stewart}. However, it is widely recognized that the majority of these solutions are complex numbers and can be eliminated by utilizing an initial guess in proximity to the true solution \cite{nguyen1991efficient,safeena2022survey}.

The design of the proposed string-based PTM is illustrated in Fig. \ref{fig:ptm_module}. This module incorporates six string potentiometers (M150, TE Connectivity, Schaffhausen, Switzerland), each with a high resolution of $0.2$ mm. These potentiometers are designed to handle a maximum stroke of $38$ mm and weigh $14$ g. To maintain appropriate tension, each potentiometer employs a torque spring, exerting a force of approximately $1.1$ N. These potentiometers are mounted on the base adapter, which is affixed to the end effector of the 6-DoF robotic manipulator, as shown in Fig. \ref{fig:PTM_Base}. The ends of the strings are securely anchored at six distinct positions on a custom-made teeth brace, as drawn in Fig. \ref{fig:PTM_Anchor}. The coordinate transformation between the base and the anchor, denoted as $_{A}^{B}\textbf{T}$, is determined during the surgical procedure by measuring the lengths of the six strings ($l_i, i \in \{1,2,\cdots,6\}$) and solving for the forward kinematics in real time.

To accommodate the restricted maximum stroke of the string potentiometers, which is only 38 mm, an extension string has been integrated. This extension guarantees that when the endodontic file, held by the dental handpiece, is inserted into the target root canal, the string potentiometer can operate throughout its entire stroke range. It's worth noting that before employing the string potentiometers, a calibration procedure is conducted to establish the conversion between the output voltage and the actual string length.

\begin{table}[!t]
    \centering
     \caption{PTM string base and anchor positions, in the coordinate frame $\{A\}$ and  $\{B\}$, respectively.}
    \begin{tabular}{crrr} 
    \hline \hline
    \textbf{Joint}   & x [mm]  &y [mm] &z [mm]      \\
    \hline  
    $\mathbf{a}_{1}$   &$-1.51$	&$-9.55$  &$-10.36$		           \\
    $\mathbf{a}_{2}$   &$-0.17$	&$-3.64$  &$-5.76$ 	    		   \\
    $\mathbf{a}_{3}$   &$1.24$	&$-16.37$ &$-6.91$		           \\
    $\mathbf{a}_{4}$   &$-1.51$	&$8.10$    &$-8.47$                \\
    $\mathbf{a}_{5}$   &$-0.06$	&$12.38$  &$-4.20$	               \\
    $\mathbf{a}_{6}$   &$-0.41$	&$4.25$   &$-4.70$             	   \\
    
    \hline 
     $\mathbf{b}_{1}$   	 &$29.73$	&$34.78$  &$12.41$ 	          \\
     $\mathbf{b}_{2}$	 &$29.73$	&$35.22$  &$45.59$		      \\
     $\mathbf{b}_{3}$	 &$29.73$	&$44.78$  &$31.41$		      \\
     $\mathbf{b}_{4}$	 &$29.73$	&$-35.22$  &$12.41$           \\
     $\mathbf{b}_{5}$	 &$29.73$	&$-45.22$  &$31.41$	          \\
     $\mathbf{b}_{6}$	 &$29.73$	&$-34.78$  &$45.59$           \\
    \hline\hline
    \end{tabular}
    \label{tab:PTM_Joint_Position}
    \vspace{-5mm}
\end{table}

To enhance string length sensitivity and prevent any collisions between the strings and the dental handpiece, a deliberate arrangement of string potentiometers and anchor positions has been implemented. The goal is to maximize the distance between each pair of base or anchor joints while minimizing the overall footprint of the module. For the base adapter, three string potentiometers are positioned on each side of the dental handpiece to evenly distribute the tension forces generated by the potentiometers, as illustrated in Fig. \ref{fig:PTM_Base}. On each side, the sensors are situated to maximize the area spanned by the base positions of the strings ($\mathbf{b}_i, i \in \{1,2,\cdots,6\}$), but meanwhile maintain a safe distance between the strings and the dental handpiece. A similar rationale is followed in determining the anchor positions ($\mathbf{a}_i, i \in \{1,2,\cdots,6\}$). An example of the teeth brace can be seen in Fig. \ref{fig:PTM_Anchor}. Further details regarding the joint positions are provided in Tab. \ref{tab:PTM_Joint_Position}. It is important to note that the order of anchor positions along the $y$-axis closely matches the order of the string bases. This arrangement helps minimize the risk of strings crossing over each other.


\subsection{Kinematic Computation and Integration}
The PTM estimates the relative pose between the robot and the patient, i.e., the pose vector $\mathbf{p_s} \in \mathbb{R}^6 $ that represents the linear and angular displacements of the homogeneous transformation matrix between the anchor frame $\{A\}$ and the base frame $\{B\}$. This estimation is achieved by solving the forward kinematics problem of the proposed string-based parallel kinematic mechanism. In this work, the Jacobian-based Newton-Raphson method \cite{nguyen1991efficient} is applied.  

To start with, the method solving for the inverse kinematics problem of the PTM is introduced. For any relative pose, the corresponding string length $l_i$ can be obtained by
\begin{equation}
l_i = \left \|  _{A}^{B}\textbf{R} ^A\! \mathbf{a}_{i}
+ {^\mathrm{B}\!\mathbf{t}_\mathrm{A}}
- \,^B\! \mathbf{b}_{i}\right \|_2,\:  i \in \{1,2,...,6\},
\end{equation}
where $\mathbf{a}_{i}$ and $\mathbf{b}_{i}$ denote the anchor and base joint position, respectively. The calculation of inverse kinematics at the relative pose $\mathbf{p_s}$ is denoted as $InvKine$($\mathbf{p_s}$), which returns the six string lengths, denoted as $\mathbf{l}_{\mathbf{p}} \in \mathbb{R}^6$, $\mathbf{l}_{\mathbf{p}} = \begin{bmatrix}
l_1 & l_2 & \cdots &  l_6
\end{bmatrix}^T$.

Given the measured string lengths $\mathbf{l} \in \mathbb{R}^6$, the forward kinematics is solved iteratively. In each step, the estimated string lengths $\mathbf{l}_{\mathbf{p}}$ are calculated from the inverse kinematics at the current estimated pose. Then, the string length error $\Delta\mathbf{l}$ between the measurement $\mathbf{l}$ and the estimate $\mathbf{l}_{\mathbf{p}}$ are computed and used to update the estimated pose.  The iterative solver continues until $||\Delta\mathbf{l}||_2$ is less than the predefined error tolerance, $tol$, which is set as $1 \times 10^{-10}$ mm in this work. This pose estimation algorithm is summarized as Algorithm \ref{algo:cable pose}. 
Note that $\mathbf{p}_{s0}$ denotes the initial guess, which is set as the relative pose estimated at the previous sampling time. $J(\mathbf{p}_s)$ returns the Jacobian matrix, which characterizes how the relative pose changes at $\mathbf{p}_s$ in response to variations in string lengths. This matrix is carried out using a numerical sensitivity analysis.

\begin{algorithm}[!b]
\caption{Pose Estimation Algorithm}
\label{algo:cable pose}
\begin{algorithmic}[1]
\Procedure{PoseEst}{$\mathbf{l},\ \mathbf{p}_{s0}, \ \textit{tol}$}
    \State $\mathbf{p}_s \gets \mathbf{p}_{s0}$
    \Do
        \State $\mathbf{l}_{\mathbf{p}} \gets  InvKine(\mathbf{p}_s)$
        \State $\Delta\mathbf{l} \gets \mathbf{l} - \mathbf{l}_{\mathbf{p}}$
        \State $\mathbf{p}_s \gets \mathbf{p}_s + (J(\mathbf{p}_s))^{-1} \cdot \Delta\mathbf{l}$
    \doWhile{$||\Delta\mathbf{l}||_2 > \textit{tol}$}
    \State \Return $\mathbf{p}_s$
\EndProcedure
\end{algorithmic}
\end{algorithm}

Last but not least, the estimated pose vector $\mathbf{p}_s$ in the PTM base frame $\{B\}$ is transformed to the file frame $\{F\}$ for patient tracking control, which will be elaborated in Section \ref{sec:Intra-guid}.


\subsection{Kinematic Performance Analysis}
The kinematic performance of the proposed string-based PTM is analyzed in this subsection. First, the error sensitivity of PTM estimation using the proposed configuration is analyzed and compared with that of the other two configurations. Second, we examine the workspace of the proposed PTM. Last, we investigate the presence of any singularities, mechanical interference, or string-crossing issues during the movement of the endodontic file within the designated cylinder workspace. All the simulations are done with the kinematic model of the DentiBot, including the PTM joint positions listed in Tab. \ref{tab:PTM_Joint_Position}.

\begin{figure}
     \centering
     \includegraphics[width=8.5cm]{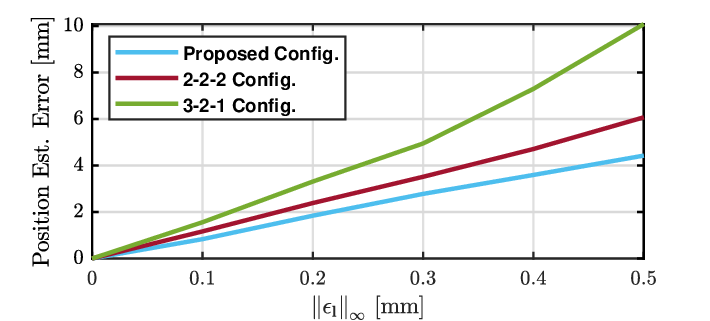}
     \caption{Maximum position estimation errors in the Monte Carlo simulation ($N$=$10,000$) under disturbed string length measurements. Different string-based parallel kinematic mechanisms are tested and compared, namely the 3-2-1 type, the 2-2-2 type, and the proposed configuration. The results indicate that the proposed configuration yields the lowest estimation error. Note that $\left \| \mathbf{\epsilon_{l}} \right \|_\infty$ denotes the maximum measuring error of string lengths. In practice, it is around $0.2$ mm.}
     \label{fig: noise error}
 \end{figure}

A Monte Carlo simulation is conducted to assess the error sensitivity of pose estimation when employing different string-based parallel kinematic mechanisms. We compared three distinct types: the 3-2-1 type, the 2-2-2 type, and the proposed configuration. In each case, the maximum position estimation error is found across $N$=$10,000$ simulations, considering randomly perturbed string length measurements. Note that the actual pose was set at the origin, with an initial guess positioned 20 mm away from it. The disturbance $\mathbf{\epsilon_{l}} \in \mathbb{R}^6$ is generated as a uniformly distributed random number with the maximum ranging from $0$ to $0.5$ mm. The simulation results, as shown in Fig. \ref{fig: noise error}, indicate that the proposed method has less estimation error than the other two types. In practice, the maximum measuring error of string lengths is around $0.2$ mm, which results in a maximum estimation error of $1.84$ mm and mean error of $0.57$ mm. This result satisfies the requirement for accurate patient tracking (S3).

\begin{figure}
         \centering
        \subfloat[PTM workspace analysis]{
        \centering
        \includegraphics[width=8.5cm]{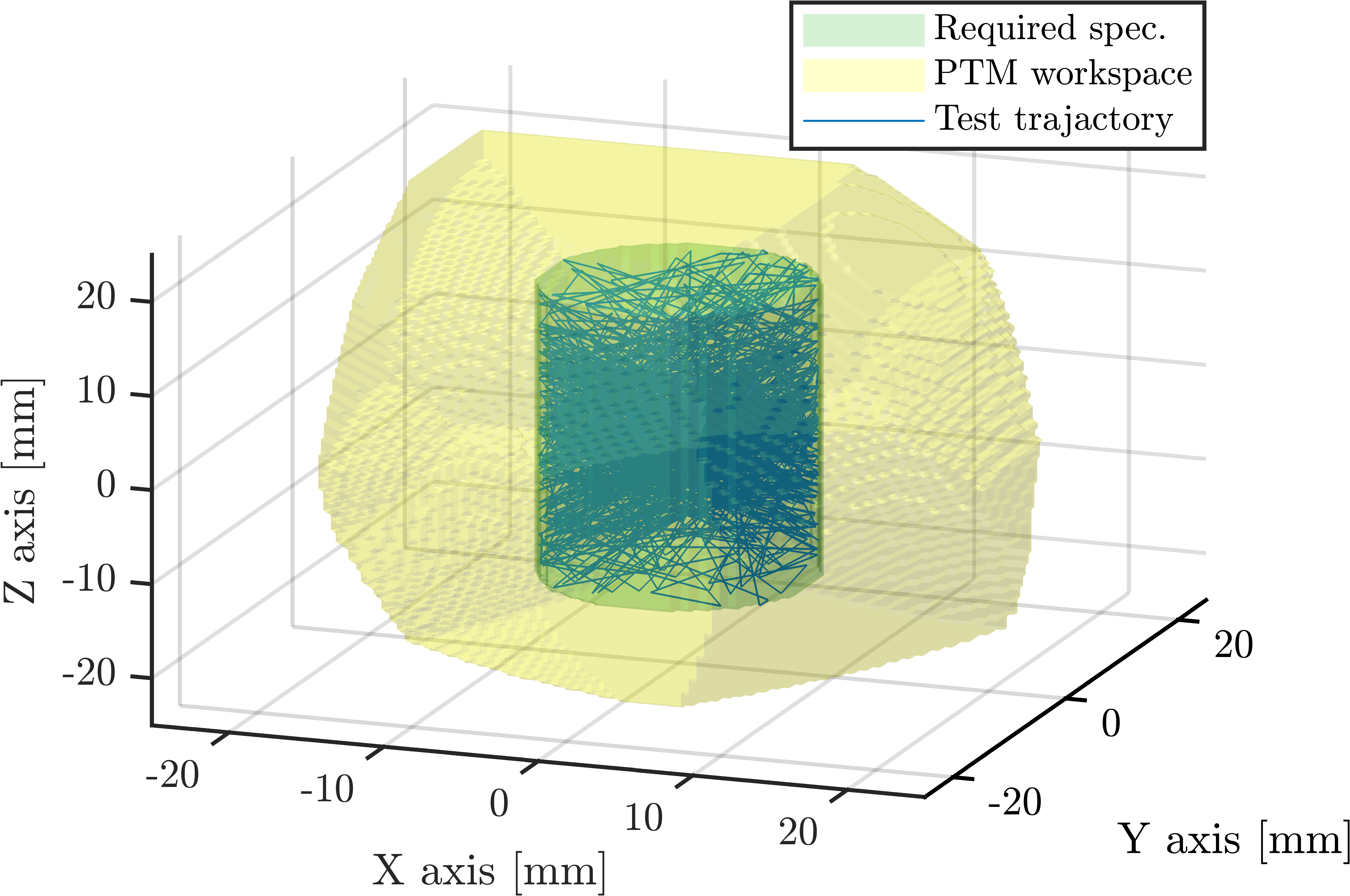}
        \label{fig:work_space_analysis}
        }\quad
        \subfloat[String-crossing and mechanical interference analysis]{
        \centering
        \includegraphics[width=8.5cm]{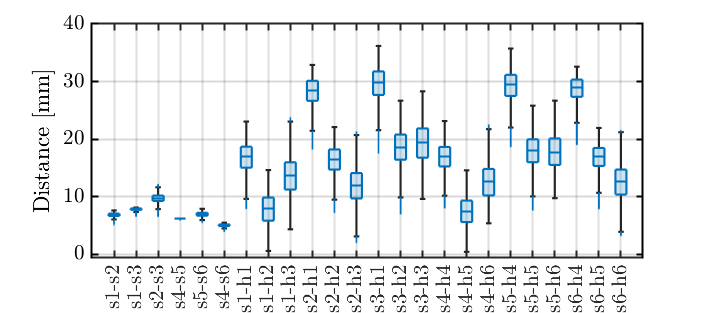}
        \label{fig:safety_distance}
        }
        \caption{Kinematic performance of the PTM. (a) The required cylindrical workspace is completely confined within the volume where the PTM attains full dexterity without exceeding the maximum stroke of the string potentiometers. (b) The distances between the six strings ($\mathfrak{s}_i, i \in \{1,2,\cdots,6\}$) and the six side edges of the hexagonal prism enclosing the dental handpiece ($\mathfrak{h}_i, i \in \{1,2,\cdots,6\}$) are computed as the endodontic file follows a random test trajectory outlined in (a). The box chart illustrating these recorded distances demonstrates the absence of string crossings or collisions between the strings and the handpiece. Notably, the minimum distance of 0.4 mm (associated with pair $\mathfrak{s}_4$-$\mathfrak{h}_5$) is observed solely in an extreme roll angle of the handpiece.}
        \label{fig:cable_workspace_analy}
\end{figure}

The analysis of PTM workspace aims to ascertain its suitability for facilitating unrestricted maneuvering of the endodontic file within the prescribed cylindrical workspace. Given that the 6-DoF robotic manipulator's workspace inherently exceeds the necessary cylindrical workspace, constraints are imposed on the PTM workspace based on the $38$ mm stroke limitation of the string potentiometers. It is assumed that the PTM is initialized with the endodontic file's tip positioned at the central point of the cylindrical workspace. In this specified pose, each string length can be extended or shortened by a maximum of $19$ mm. The workspace analysis is conducted within a cubic volume with a $40$ mm width surrounding the cylindrical workspace, employing a $0.5$ mm resolution. Dexterity, defined as the ratio of valid orientations within a $\pm5$-degree range along the roll and pitch angles, is calculated for each test point within the cube. Test points are deemed part of the PTM workspace only if they satisfy two conditions: (1) the stroke of the string potentiometers remains within the defined constraint and (2) full dexterity can be achieved at that specific point. The outcome of this workspace analysis is illustrated in Fig. \ref{fig:work_space_analysis}, where the green and yellow volumes represent the required cylindrical workspace and the PTM workspace, respectively. Importantly, the required cylindrical workspace is entirely encompassed within the region where the PTM achieves full dexterity without exceeding the maximum stroke of the string potentiometers.

Subsequently, the assessment of singularities or multiple solutions within the required cylindrical workspace for PTM estimation is conducted. In this analysis, the tip of the endodontic file undergoes random movements within the cylindrical workspace over a duration of $5000$ seconds, sampled at intervals of $0.01$ seconds. The maximum velocity of the DentiBot is constrained to $2.5$ mm/s, reflecting the maximum speed of patient movement. The test trajectory is depicted in Fig. \ref{fig:work_space_analysis}. Initially, both the starting point and the initial estimate $p_{s0}$ are positioned at the center of the cylindrical workspace. Then, the estimated pose from the preceding timestamp serves as the initial estimate for the subsequent timestamp. Consequently, no instances of singularity or multiple solutions are observed throughout the analysis. The maximum position and orientation error along the test trajectory are $1.77\times 10^{-12}$ mm and $1.11\times 10^{-12}$ degrees, respectively.

Last, an analysis is conducted to confirm there is no string-crossing and mechanical interference when applying the PTM along with the DentiBot. When analyzing the PTM estimation error along the test trajectory in the previous assessment, the distances between the six strings ($\mathfrak{s}_i, i \in \{1,2,\cdots,6\}$) and the six side edges of the hexagonal prism enclosing the dental handpiece ($\mathfrak{h}_i, i \in \{1,2,\cdots,6\}$) are also computed. Note that due to the symmetric design of the PTM, $\mathfrak{h}_1$, $\mathfrak{h}_2$, and $\mathfrak{h}_3$ are closer to $\mathfrak{s}_1$, $\mathfrak{s}_2$, and $\mathfrak{s}_3$, while $\mathfrak{h}_4$, $\mathfrak{h}_5$, and $\mathfrak{h}_6$ are closer to $\mathfrak{s}_4$, $\mathfrak{s}_5$, and $\mathfrak{s}_6$. Out of a potential $66$ interference pairs, we assess $24$, as some pairs, like $\mathfrak{s}_1$-$\mathfrak{h}_4$, are unlikely to experience interference. The results, depicted in Fig. \ref{fig:safety_distance}, demonstrate that the distances between strings consistently exceed $5$ mm. The narrowest separation between strings and the dental handpiece is $0.4$ mm, observed in the case of pair $\mathfrak{s}_4$-$\mathfrak{h}_5$ during an extreme roll angle of the handpiece. This analysis confirms the absence of string-crossing and mechanical interference when employing the PTM.

\section{Intraoperative Guidance and Control}
\label{sec:Intra-guid}
In this section, we present the proposed methods for force guidance and control during the cleaning and shaping step in RCT. The deflection of the endodontic file is estimated and compensated. The 6-DoF hybrid position/force controller is designed to minimize the contact force between the endodontic file and the root canal while simultaneously maintaining the relative pose between the robot and the patient. Furthermore, we have achieved full automation of the workflow for the cleaning and shaping step.


\subsection{Endodontic File Flexibility Model}

The distinctive feature of endodontic files lies in their flexibility, setting robot-assisted RCT apart from automated dental implant surgery. This flexibility, however, introduces an additional displacement when external forces are applied to the file, necessitating careful compensation. In this study, we compensate for the displacement resulting from file bending by implementing a virtual spring force proportional to the estimated amount of deflection. The method for estimating file deflection is introduced in this subsection.

\begin{figure}
  \centering
        \includegraphics[width=7cm]{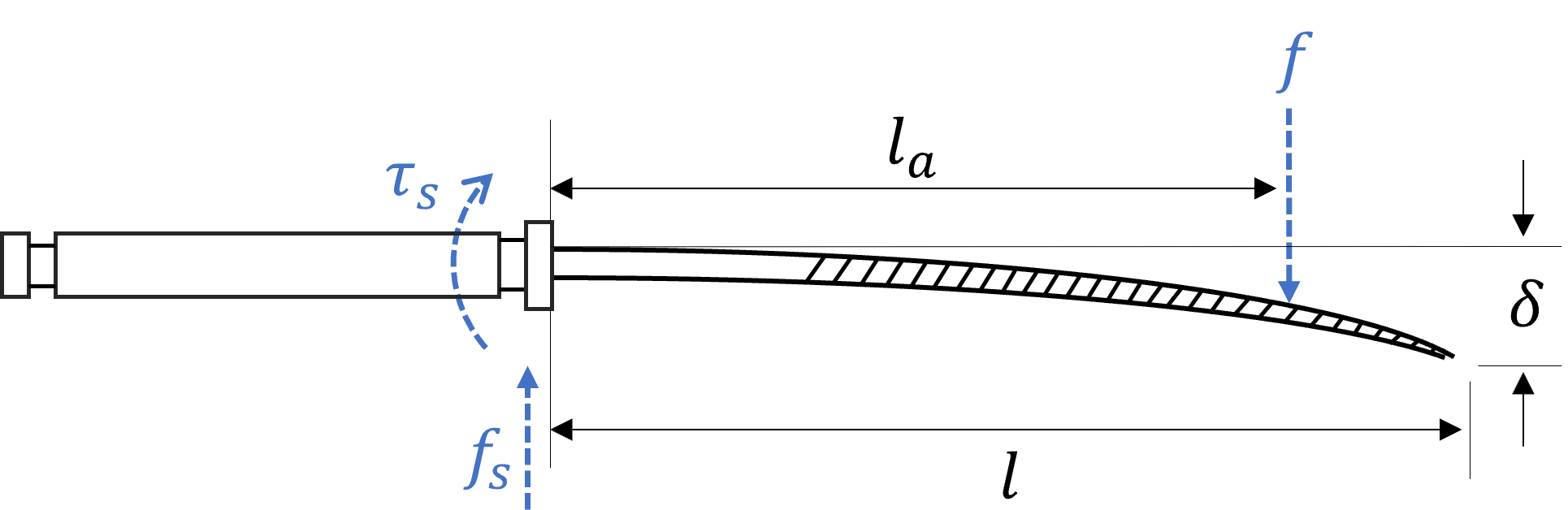}
        \caption{Model of endodontic file flexibility.     
        To estimate the deflection $\delta$ at the tip of an endodontic file in response to the resultant external force $f$, one can rely on force/torque sensing data ($f_s$ and $\tau_s$). It is worth noting that the force/torque sensing data can also be employed to determine the effective leverage length $l_a$, which is an essential term when applying the beam deflection formula (Eq. \ref{eq:FileDeflectEsti}).}
        \label{fig: cantilever_assumption}
        \vspace{-3mm}
\end{figure}

The endodontic file is considered as a beam fixed at one end on the dental handpiece and subjected to external forces from multiple directions on the other end. Since the deflection along the axial axis is negligible, only the resultant radial force $f$ is considered, as illustrated in Fig. \ref{fig: cantilever_assumption}. Note that $f$ represents the summation of external forces and surface pressure inside the root canal as a single resultant force. The absolute value of $f$ is identical to the sensed force (by abuse of notation, denoted as $f_s$ here) to yield the net force equals zero. As a result, the effective leverage length $l_{a}$ can be derived as 
\begin{equation}
    l_{a}= \frac{\mathbf{\tau_{s}}}{f} = \frac{\mathbf{\tau_{s}}}{f_{s}}, 
\end{equation}
where $\tau_s$ denotes the sensed torque.

Next, the displacement of the file tip caused by the resultant external force $f$, denoted as the maximum deflection $\delta$, can be obtained from the beam deflection formula \cite{gibson2016principles}: 
\begin{equation}
\delta = \frac{f_{s}l_{a}^{2}}{6EI}(3l-l_{a}).
\label{eq:FileDeflectEsti}
\end{equation}
Note that $E$ represents Young's modulus, which is around $80$ Gpa for NiTi files. $I$ denotes the cross-sectional area moment of inertia of endodontic files. Given that the diameter of endodontic files varies from $0.3$ mm at the tip to $1.2$ mm at the base, we consider half of the maximum diameter, i.e., $0.6$ mm, for inertia calculations. This model (Eq. \ref{eq:FileDeflectEsti}) can be applied separately to the x- and y-axes of the endodontic file frame $\{F\}$. 

The compensation for the estimated file deflection involves applying a virtual spring force. This virtual spring force will be integrated with the 6-DoF hybrid position/force control, which will be further explained in the subsequent subsection. It is worth noting that the estimated leverage length $l_{a}$ may exceed the length of the file due to measurement noises, particularly when only the file tip is in contact or when the applied external force is relatively small. To tackle this issue, we establish an upper limit for $l_a$, which is set equal to $l$, the known length of each endodontic file. Additionally, the file flexibility compensator is activated only when the measured force along the x- or y-axis exceeds a predefined threshold, which we have set at $0.03$ N in our system.

\subsection{6-DoF Hybrid Position/Force Control}

In order to minimize the contact force between the endodontic file and the root canal while simultaneously maintaining the relative pose between the robot and the patient, a hybrid position/force control strategy is proposed for the DentiBot. The position control utilizes the pose estimation $\mathbf{p}_s$ from the PTM to track the patient's motion. The force control applies the force/torque measurement $\boldsymbol{\xi}_s \in \mathbb{R}^6$,
\begin{equation}
    \boldsymbol{\xi}_s = \begin{bmatrix}
    \boldsymbol{f}_s\\ 
    \boldsymbol{\tau}_s
    \end{bmatrix},
\end{equation}  
to adjust the file insertion path and prevent file failure or ledging. By combining these two control strategies, the hybrid position/force controller maintains a delicate balance between tracking accuracy and minimizing contact forces in real-time. This ensures that the DentiBot remains precise and effective throughout the cleaning and shaping process.

\begin{figure}[!t]
\centering
    \includegraphics[width=8.5cm]{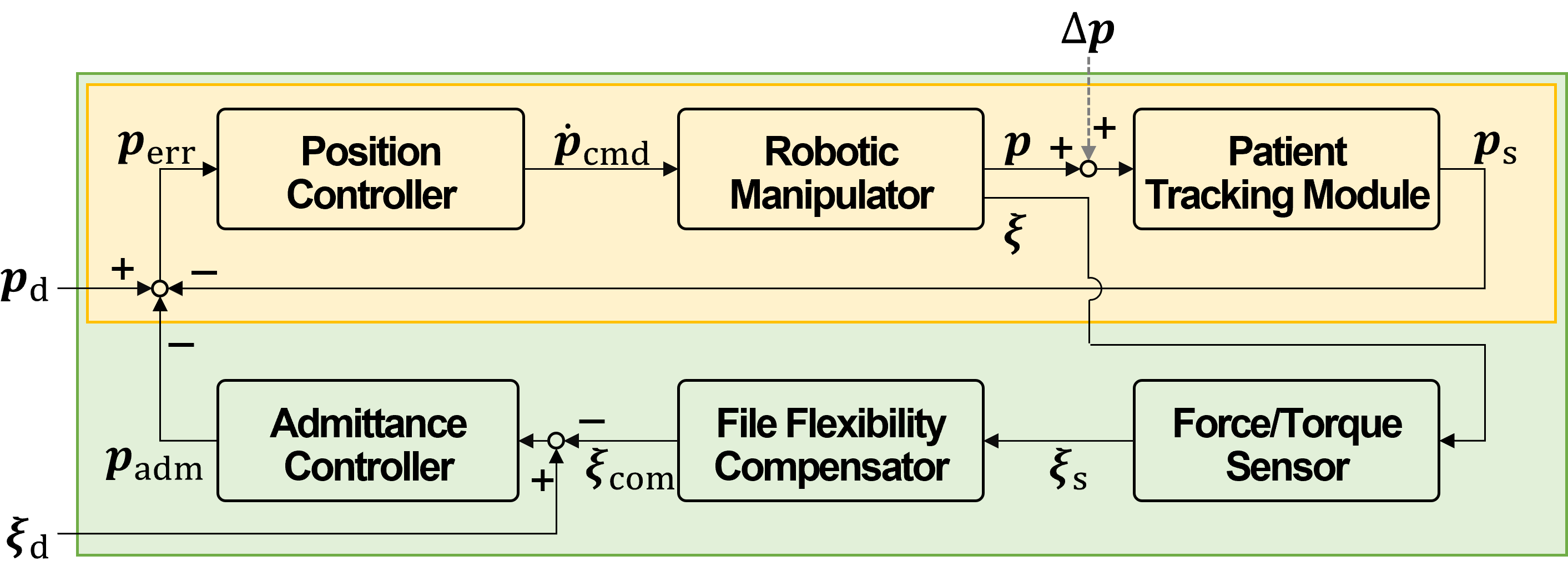}
\caption{Block diagram of the proposed 6-DoF hybrid position/force control. In the inner-loop position control (yellow box), the robotic manipulator receives velocity commands $\dot{\mathbf{p}}_{cmd}$ and adjusts its pose (relative to the patient) to reach the desired value ${\mathbf{p}}_{d}$. Here, ${\Delta\mathbf{p}}$ denotes the patient movement, and ${\mathbf{p}}_{s}$ represents the relative pose measured from the PTM. In the outer-loop force control (green box), the interaction force between the endodontic file and root canal, i.e., $\mathbf{\xi}_{s}$ measured from the force/torque sensor, is transmitted to the file flexibility compensator and the admittance controller. These components generate the pose adjustment $\mathbf{p}_{adm}$ to track the desired interaction force $\mathbf{\xi}_d$.}
\label{fig: overall structure}
\vspace{-3mm}
\end{figure}

The control block diagram for the DentiBot is shown in Fig. \ref{fig: overall structure}. The controller comprises of an inner and an outer loop, which work together to track patient movement and adjust file insertion path, respectively. The inner loop, running at the sampling rate of $100$ Hz, implements a closed-loop position control to regulate the relative pose between the robot and the patient, i.e., $\mathbf{p}_{s}$ measured from the PTM, to the initial pose $\mathbf{p}_{d}$. The outer loop, running at the sampling rate of $20$ Hz, generates the force-guided path correction $\mathbf{p}_\mathrm{adm}$ based on the force/torque measurement $\boldsymbol{\xi}_s$, which then modifies the initial pose $\mathbf{p}_{d}$. This hybrid configuration is also called ``implicit force control'' in literature \cite{winkler2015implicit}. Note that all the signals are represented in the file frame $\{F\}$.

In the inner control loop, the difference between $\mathbf{p}_{d}$ (or $\mathbf{p}_{d}-\mathbf{p}_{adm}$ if $\mathbf{p}_{adm}$ is a non-zero vector) and $\mathbf{p}_{s}$, denoted as $\mathbf{p}_\mathrm{err}$, is fed into a proportional-derivative (PD) controller, which generates the velocity command $\mathbf{\dot{p}}_\mathrm{cmd}$ to the 6-DoF robotic manipulator: 
\begin{equation}
    \mathbf{\dot{p}}_\mathrm{cmd}(k)= C_\mathrm{PD}(z)  \mathbf{p}_\mathrm{err}(k),
\label{eq:PD}
\end{equation}
where
\begin{equation}
    \mathbf{p}_\mathrm{err}(k) = \mathbf{p}_\mathrm{d}(k) - \mathbf{p}_\mathrm{adm}(k) - \mathbf{p}_\mathrm{s}(k).
\end{equation}
Note that $k$ is the discrete time index. $C_\mathrm{PD}(z)$ is the six-dimensional PD position controller with the proportional gain $k_p$ and the derivative gain $k_d$ for each linear or angular axis. When the patient moves, i.e., $\Delta\mathbf{p}\neq \mathbf{0}$, the PTM estimates the $\Delta\mathbf{p}$ along with the current robot pose $\mathbf{p}$, resulting in none-zero $\mathbf{p}_\mathrm{err}$. The use of a PD position controller is advantageous as it provides closed-loop stability, i.e., asymptotically convergence of $\mathbf{p}_\mathrm{err}$, and ease of parameter tuning, while also preventing overshoot or oscillations within the narrow space of the root canal. 

As for the outer control loop, the measured contact force and torque $\boldsymbol{\xi_{s}}$ are sent to the file flexibility compensator, which estimates the file deflection along the x- and y-axis, denoted as $\delta_{x}$ and $\delta_{y}$, respectively, using Eq. \ref{eq:FileDeflectEsti}. Next, the compensation for file deflection is achieved by introducing virtual spring forces applied to the original force/torque measurements: 
\begin{equation}
    \boldsymbol{\xi}_\mathrm{com}(k) = \boldsymbol{\xi}_\mathrm{s}(k)+ \boldsymbol{\xi}_\mathrm{flx}(k),
\end{equation}
where
\begin{equation}
    \boldsymbol{\xi}_\mathrm{flx}(k) = [k_{f}\delta_{x}(k) \  k_{f}\delta_{y}(k)  \ 0 \  0 \  0 \ 0]^{T}.
    \label{eq:fileflexcomp}
\end{equation}
Note that $k_{f}$ denotes the virtual spring constant, which is a design parameter in the file flexibility compensator. The inclusion of extra forces during file flexibility compensation allows the admittance controller, the next function block, to generate a larger path adjustment, thereby diminishing file deflection.

The DentiBot utilizes an admittance controller to ensure appropriate response to contact force and torque \cite{cheng2022force}. Admittance control has been widely adopted to improve the safety during human-robot collaboration \cite{augugliaro2013admittance}. Generally, admittance control enables a robot to mimic the behavior of a second-order mass-damper system \cite{350927}. In the context of autonomous cleaning and shaping in RCT, the desired force/torque, denoted as $\boldsymbol{\xi}_\mathrm{d}$, consists of zero values for all elements except the z-axis and $\theta$-axis (file rotation around the z-axis). The endodontic file's insertion motion is driven by a positive desired contact force along the z-axis. Conversely, no control is exerted on the $\theta$-axis to prevent the interactive torque generated by file rotation and drilling from affecting the DentiBot's movement. Apart from these two axes, the contact force/torque should remain at zero.

Applying admittance control, the force-guided path correction $\mathbf{p}_\mathrm{adm}$ is calculated by  
\begin{equation}
\label{eq:adm_mb}
\begin{split}
\mathbf{p}_\mathrm{adm}(k)
=&  C_{adm}(z) 
\left(\boldsymbol{\xi}_\mathrm{d}(k) - \boldsymbol{\xi}_\mathrm{com}(k)
\right),
\end{split}
\end{equation}
where
\begin{equation}
C_\mathrm{adm}(z) =diag \left( \frac{k_aT(z+1)^{2}}{4m_a(z-1)^{2}+2b_{a}T(z^2-1)}\right).
\label{eq:adm}
\end{equation}
Note that the mass coefficient $m_a$, damper coefficient $b_a$, and constant gain $k_a$ are design parameters in the admittance controller. These parameters can vary for each linear axis ($x$, $y$, $z$) and rotary axis ($\phi$, $\psi$, $\theta$). The sampling period of the outer control loop is denoted as $T$, which is set as $0.05$ s in our system. 

The hybrid position/force control strategy is utilized specifically during the cleaning and shaping phase of RCT. Prior to this step or in risky situations, only admittance control is employed, allowing the dentist to manually handle and maneuver the dental handpiece. When operating under pure admittance control, the PD position controller is bypassed, meaning that the admittance controller directly generates the velocity command $\mathbf{\dot{p}}_\mathrm{cmd}$ for the 6-DoF robotic manipulator. The transition between each of these automated steps will be further elaborated in the subsequent subsection.



\subsection{Automation of Surgical Tasks}

The typical workflow of robot-assisted root canal cleaning and shaping, utilizing the DentiBot system, is outlined as follows. After the dentist creates an access open to the pulp chamber and root canal, the autonomous cleaning and shaping procedure commences. The admittance control is activated, enabling the dentist to manipulate the dental handpiece, attached to the DentiBot, and align the endodontic file with the target root canal. Subsequently, the PTM establishes a connection between the robot and the patient, registering the initial relative pose ${\mathbf{p}}_{d}$. The endodontic file is then automatically inserted into the root canal using the hybrid position/force control. Once inserted, the cleaning and shaping of the root canal is initiated.

\begin{figure}[!t]
  \centering
        \includegraphics[width=8.5cm]{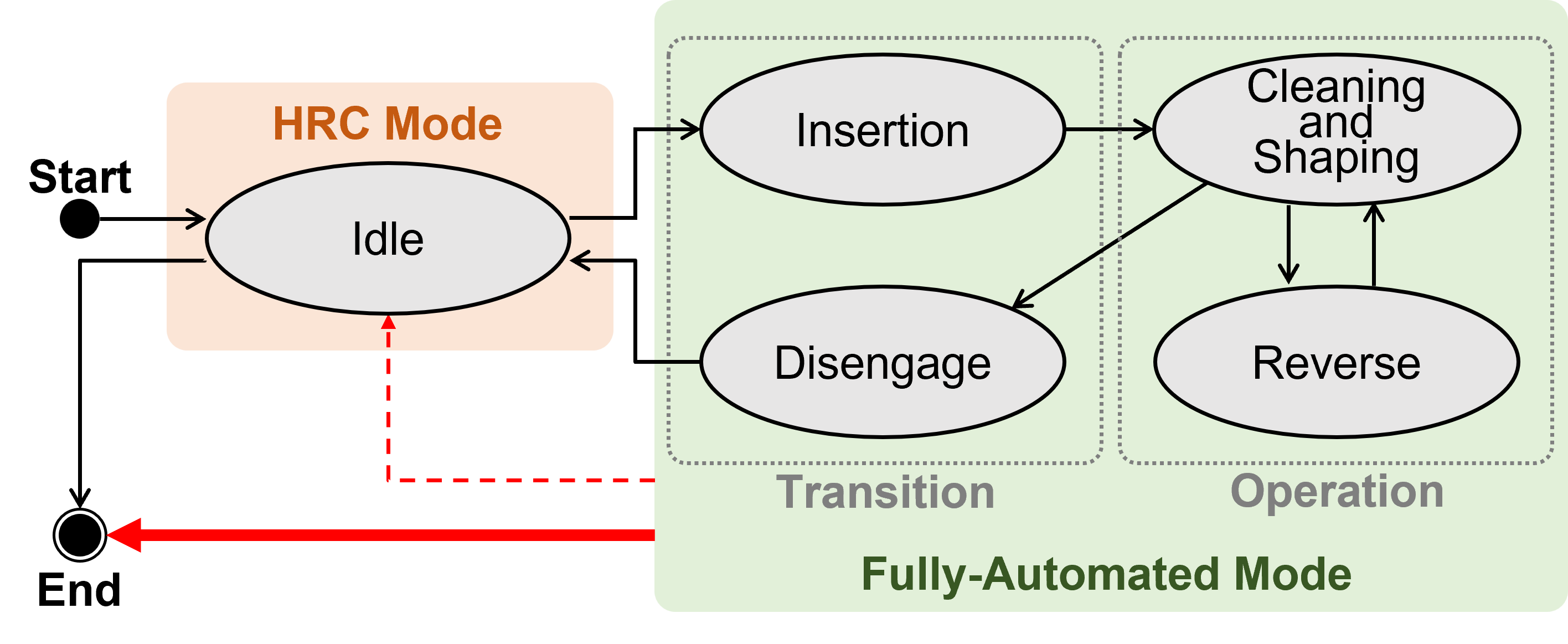}
        \caption{Statechart diagram for autonomous endodontic cleaning and shaping. First. the dentist guides the robot to the infected pulp while in the Idle state under the human-robot collaboration (HRC) mode. Next, the DentiBot enters the fully-automated mode. The endodontic file is inserted until its tip reaches the open of the root canal. Once the file is in place, cleaning and shaping begin. The file will engage until the tip reaches the apex of the root canal (i.e., the procedure is complete), or reverse if the axial torque reaches the upper limit. In case of unexpected events, the DentiBot will enter the HRC mode (red dash line) and wait for troubleshooting, or freeze the motion immediately (red solid line).}
        \label{fig:finite_machine}
        \vspace{-3mm}
\end{figure}

In the cleaning and shaping stage, the endodontic file remains engaged until its tip reaches the apex of the root canal, indicating the completion of the instrumentation. To prevent fracturing, the file automatically reverses if the axial torque exceeds the upper limit. Once the file reaches the apex, it is withdrawn from the root canal. At this point, the dentist can choose to either replace the endodontic file, typically with a larger diameter, and repeat the procedure, or decide to conclude the treatment. This workflow is modeled as a finite state machine to facilitate surgical automation. As illustrated in the statechart diagram (Fig. \ref{fig:finite_machine}), there are totally five states: Idle, Insertion, Cleaning and Shaping, Reverse, and Disengage. The specific actions and transitions associated with each state are described below.

\subsubsection{Idle} 
In the Idle state, admittance control is enabled to respond to external forces and facilitate human-robot collaboration (HRC mode). The position controller and file flexibility compensator remain deactivated. Upon the dentist positioning the dental handpiece and connecting the PTM, the system transitions to the Insertion state. It is important to note that the initial relative pose ${\mathbf{p}}_{d}$ is recorded upon exiting the Idle state.

\subsubsection{Insertion} 
Starting from the Insertion state, the DentiBot switches to the fully-automated mode. It activates the hybrid position/force control. A non-zero desired force along the z-axis, set to $0.4$ N, is defined in $\boldsymbol{\xi}_\mathrm{d}$. This allows the DentiBot to gradually insert the endodontic file into the target root canal. The DentiBot continues the insertion process until the desired contact force is detected or the insertion reaches the same depth as in the previous treatment. At this point, the DentiBot transitions to the next state for cleaning and shaping.

\subsubsection{Cleaning and Shaping} 
In this state, the DentiBot autonomously performs the cleaning and shaping of the root canal. The hybrid position/force control remains activated, and the endodontic file begins to rotate at the speed of $150$ rpm. The desired force along the z-axis is gradually increased from $0.4$ N to $0.6$ N, $0.8$ N, and $1.0$ N if the axial torque $\tau_z$ does not exceed the threshold of $8$ mN$\cdot$m for at least $15$ seconds during the cleaning and shaping procedure. However, if the torque $\tau_z$ exceeds the threshold, the DentiBot transitions to the Reverse state. In the Cleaning and Shaping state, the dentist assumes the responsibility of determining whether the file tip has reached the apex or if the root canal needs to be flushed with water to remove debris. If either condition is met, the system enters the Disengage state.

\subsubsection{Reverse}
This state is implemented to offer a protective mechanism during the cleaning and shaping process when the sensed axial torque reaches its limit. In this state, the rotation of the endodontic file is reversed, operating at the commanded speed of $250$ rpm. This reversal of rotation helps release the torque and minimize the risk of file fracture. Moreover, by reversing the file, the file is slightly withdrawn. Any remaining debris inside the root canal can then be removed, ensuring a clean root cavity. The system automatically transitions back to the Cleaning and Shaping state after spending one second in the Reverse state.

\subsubsection{Disengage} 
In this state, the DentiBot initiates a reverse rotation of the file and begins pulling it out. The withdrawal process is realized by assigning a negative desired force of $-0.8$ N along the z-axis for the admittance controller. The system transitions from the Disengage state to the Idle state once the file insertion depth returns to zero, indicating that the file has been completely extracted.

After the endodontic file is successfully extracted, the DentiBot resumes its operation in the HRC mode. Following a thorough diagnosis by the dentist, they can make an informed decision regarding whether to conclude the treatment or proceed with another round using either the same or a different file. Once the treatment is deemed complete, the dentist removes the braces from the patient's tooth and gently moves the robot aside before concluding the automated procedure.

In the event of unexpected occurrences, the DentiBot will revert to the HRC mode and await troubleshooting. This may happen if, for instance, the axial torque exceeds the limit even after reversing the rotation, or if the string lengths of the PTM approach their maximum limit. Additionally, if a hazardous situation such as a sudden and rapid patient movement is detected, the DentiBot system will halt its motion and promptly terminate the automated procedure. In both emergency overrides, the rotation of the file will cease as well.

\section{Visualization and User Interface}
\label{sec:Visualize}
In this section, we develop a graphical user interface (GUI) for the DentiBot system that allows dentists to supervise the robotic manipulator's motion in a virtual environment. Preoperative computed tomography (CT) images of the root canal are processed and displayed, enabling real-time progress monitoring of robot-assisted dental procedures.

\subsection{Preoperative CT Image Processing}
In the cleaning and shaping step of RCT, it is essential for dentists to measure the length of the root canal and monitor the depth of file insertion. Conventional manual surgery lacks intraoperative visual feedback for this process. To overcome this limitation, we propose using preoperative CT scans to create a 3D model of the target root canal. The pathway of the root canal is also detected. Our GUI displays the preoperative model along with real-time visualization of the file being inserted into the root canal. This intraoperative real-time feedback allows dentists to monitor the surgical progress during the robot-assisted procedure.

\begin{figure}[!t]
  \centering
        \includegraphics[width=8.5cm]{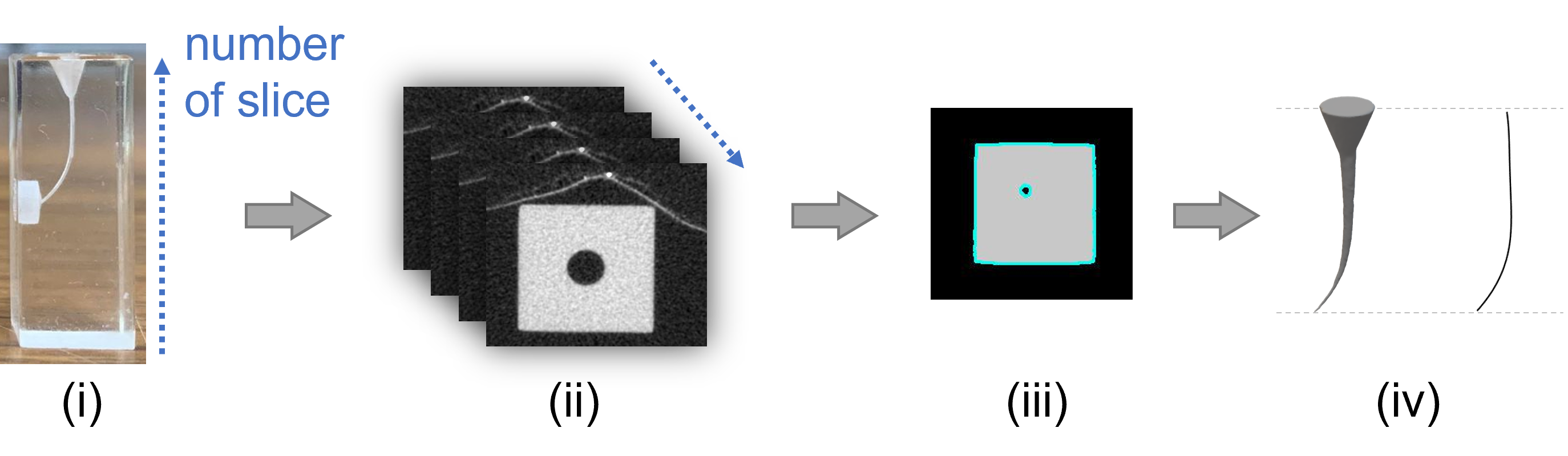}
        \caption{Procedure of root canal model reconstruction. Using the CT scans performed on the target teeth, the root canal model is constructed after stacking and segmenting the cross-sectional CT images. The endodontic file path is then obtained by finding the central path of the root canal. (i) Target teeth. (ii) Cross-sectional CT images. (iii) Segmentation for root canal. (iv) Root canal 3D model and endodontic file path. }
        \label{fig: root_path_construction}
\end{figure}

The process of reconstructing the root canal model and detecting its central path is illustrated in Fig. \ref{fig: root_path_construction}. In this study, we utilize a micro-CT scanner (U-CT, MILabs, Houten, Netherlands) to capture high-resolution CT cross-sectional images of the teeth models at a resolution of 10 $\mu$m. A sample of the root canal model and corresponding CT images is presented in Fig. \ref{fig: root_path_construction}. To segment the root canal, we convert the grayscale CT images into binary format and apply the Canny edge detection algorithm \cite{canny1983finding} to each cross-sectional image. By stacking the contours segmented from CT cross-sectional images, the root canal 3D model is reconstructed.

After retrieving the root canal 3D model, the central path of the root canal is detected by fitting a sixth-order polynomial function on the x-z plane and y-z plane, respectively: 
\begin{equation}
    \begin{bmatrix}
x_{rc}
 \\    
y_{rc}
\end{bmatrix}
=
\begin{bmatrix}
g_{1}
 \\    
h_{1}
\end{bmatrix} 
z_{rc}^6+
\begin{bmatrix}
g_{2}
 \\    
h_{2}
\end{bmatrix} 
z_{rc}^{5}
+
...
+
\begin{bmatrix}
g_{7}
 \\    
h_{7}
\end{bmatrix},
\end{equation}
where a set of $(x_{rc}, y_{rc}, z_{rc})$ satisfying the above equation denotes the central path. $g_{j}$ and $h_{j}$, $j \in \{1,2,\cdots,7\}$, represent the coefficients of the polynomial with respect to the x-axis and y-axis, respectively. The length of the central path can then be calculated, providing valuable information for dentists in selecting appropriate endodontic files for the cleaning and shaping step. It also allows them to determine if the file tip has reached the apex of the root canal.


\subsection{Graphical User Interface}
The GUI of the DentiBot system is depicted in Fig. \ref{fig:dental_gui}. It provides a visualization of the root canal 3D model and the DentiBot, which includes the robotic manipulator and dental handpiece. The virtual manipulator is capable of synchronous 6-DoF motions by receiving 6-axis joint data through the UDP protocol. Preoperative calibration establishes fixed homogeneous transformation matrices, such as $_{R}^{B}\textbf{T}$, $_{P}^{A}\textbf{T}$, and $_{F}^{R}\textbf{T}$. $_{A}^{B}\textbf{T}$ is derived from real-time PTM estimation. The relative pose between the endodontic file and the root canal model, represented as $_{P}^{F}\textbf{T}$, is thereby calculated and plotted based on 
\begin{equation}
     _{P}^{F}\textbf{T} = 
     {_{R}^{F}\textbf{T}}{_{B}^{R}\textbf{T}}
     {_{A}^{B}\textbf{T}}{_{P}^{A}\textbf{T}}.
\end{equation}

\begin{figure}[!t]
\centering
\subfloat[Graphical user interface]{%
  \includegraphics[width=8cm]{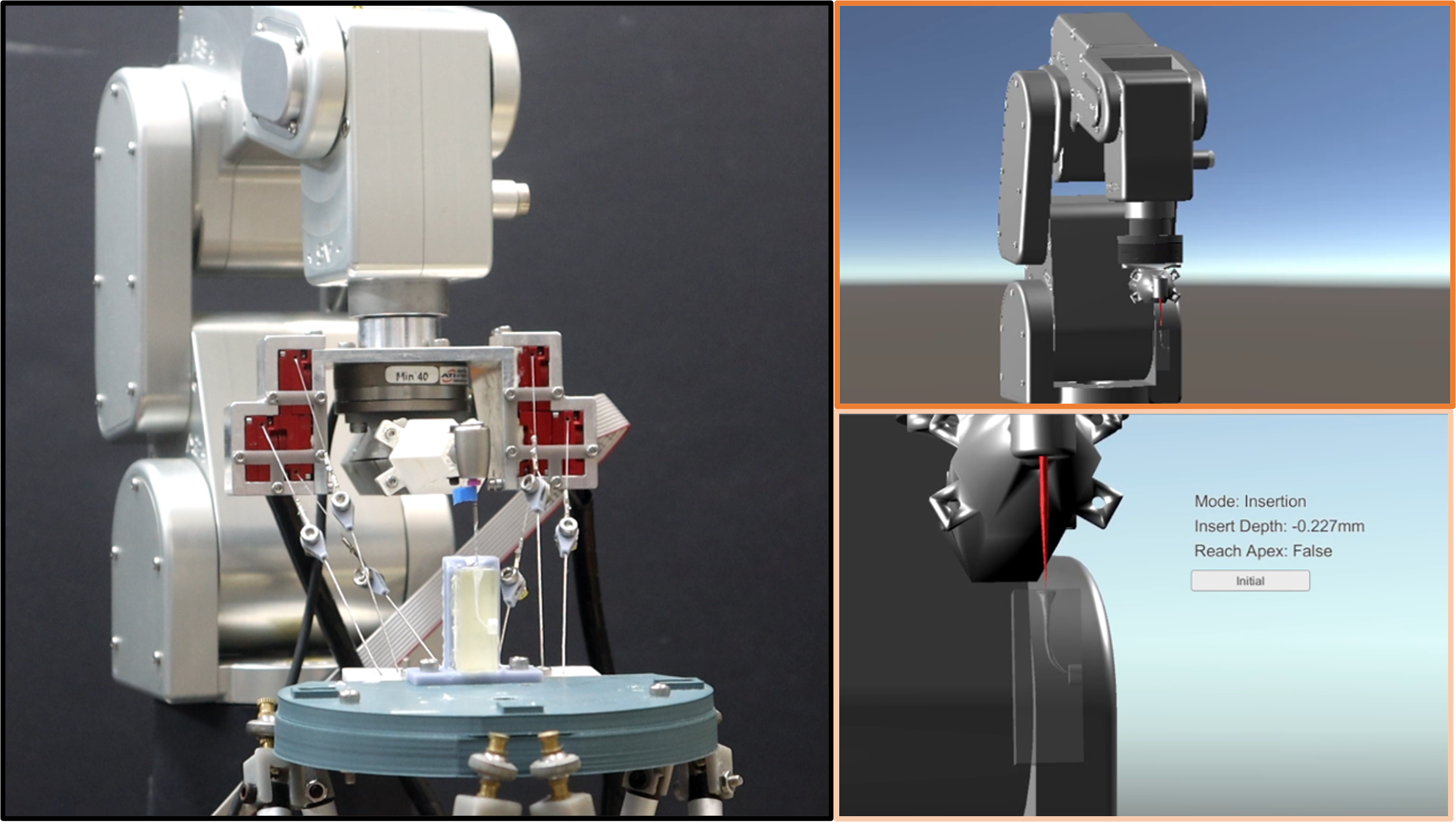}%
  \label{fig:dental_gui_all}%
}\quad
\vspace{1mm}
\subfloat[Insertion depth monitor]{%
  \includegraphics[width=8cm]{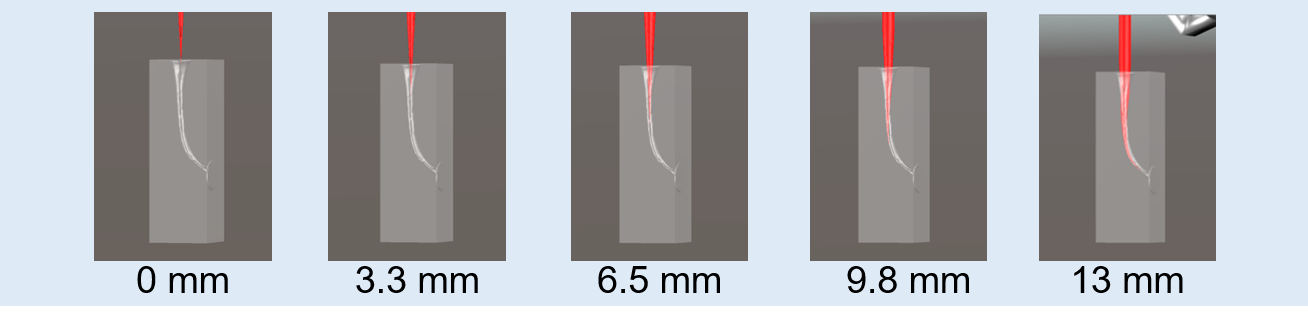}%
  \label{fig:dental_gui_insert}%
} 
\label{fig:dental_gui}
\vspace{-1mm}
\caption{Graphical user interface (GUI) of the DentiBot. The GUI demonstrates the real-time interaction between the DentiBot and the root canal, in both the actual and virtual environment. The pre-scanned CT model of the root canal allows dentists to supervise the surgical progress (e.g., file insertion depth) inside the root canal. This feature is unachievable in conventional manual surgery. (a) Graphical user interface. (b) Surgical progress demonstrated by file insertion depth. The endodontic file is marked in red. }
\label{fig:dental_gui}
\vspace{-4mm}
\end{figure}

During the autonomous cleaning and shaping process, the GUI provides dentists with visual feedback, displaying the current operation state and real-time file insertion depth. This functionality enables dentists to closely monitor the surgical progress and intervene if necessary. Fig. \ref{fig:dental_gui_insert} exemplifies the GUI interface, depicting the insertion of the endodontic file into the preoperatively reconstructed root canal. The GUI dynamically calculates and presents the real-time insertion depth, supplying dentists with crucial information throughout the procedure.

\section{Experimental Results}
\label{sec:Exp}
The evaluation of the DentiBot system involves two aspects: engineering and pre-clinical evaluations. The engineering evaluation included control parameter selection, PTM accuracy analysis, and real-time robot-patient alignment under different control schemes. In the pre-clinical evaluation, the DentiBot autonomously performed cleaning and shaping on acrylic and AA temp resin root canal models. The evaluation included assessing the success rate of the procedures as well as comparing the changes in root canal volume and length before and after the automated procedure.


\subsection{System Setup}

The DentiBot system, including the 6-DoF robotic manipulator and dental handpiece, was controlled by an NI LabVIEW Real-Time Target (Intel Core i7-3770 Processor) via the EtherCAT protocol. The Real-Time Target also received measurements from the force/torque sensor and PTM. On the other hand, the GUI was implemented on another personal computer (Intel Core i7-8700 Processor). It established communication with the Real-Time Target using the UDP protocol.

\begin{figure*}[htbp]
\centering
\subfloat[Parameter selection]{%
  \includegraphics[width=5.5cm]{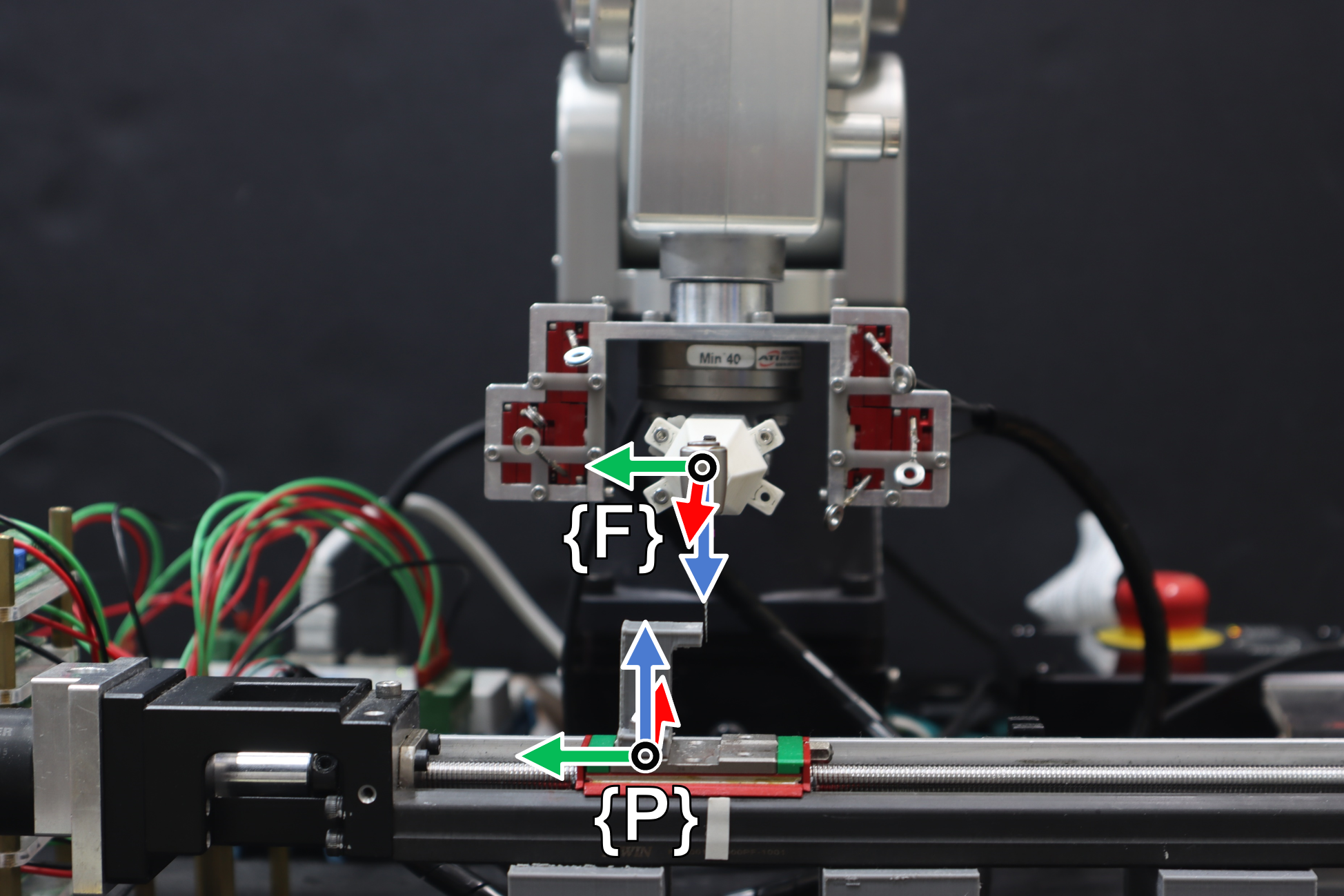}%
  \label{fig: linear_exp}%
}\quad
\subfloat[PTM evaluation]{%
  \includegraphics[width=5.5cm]{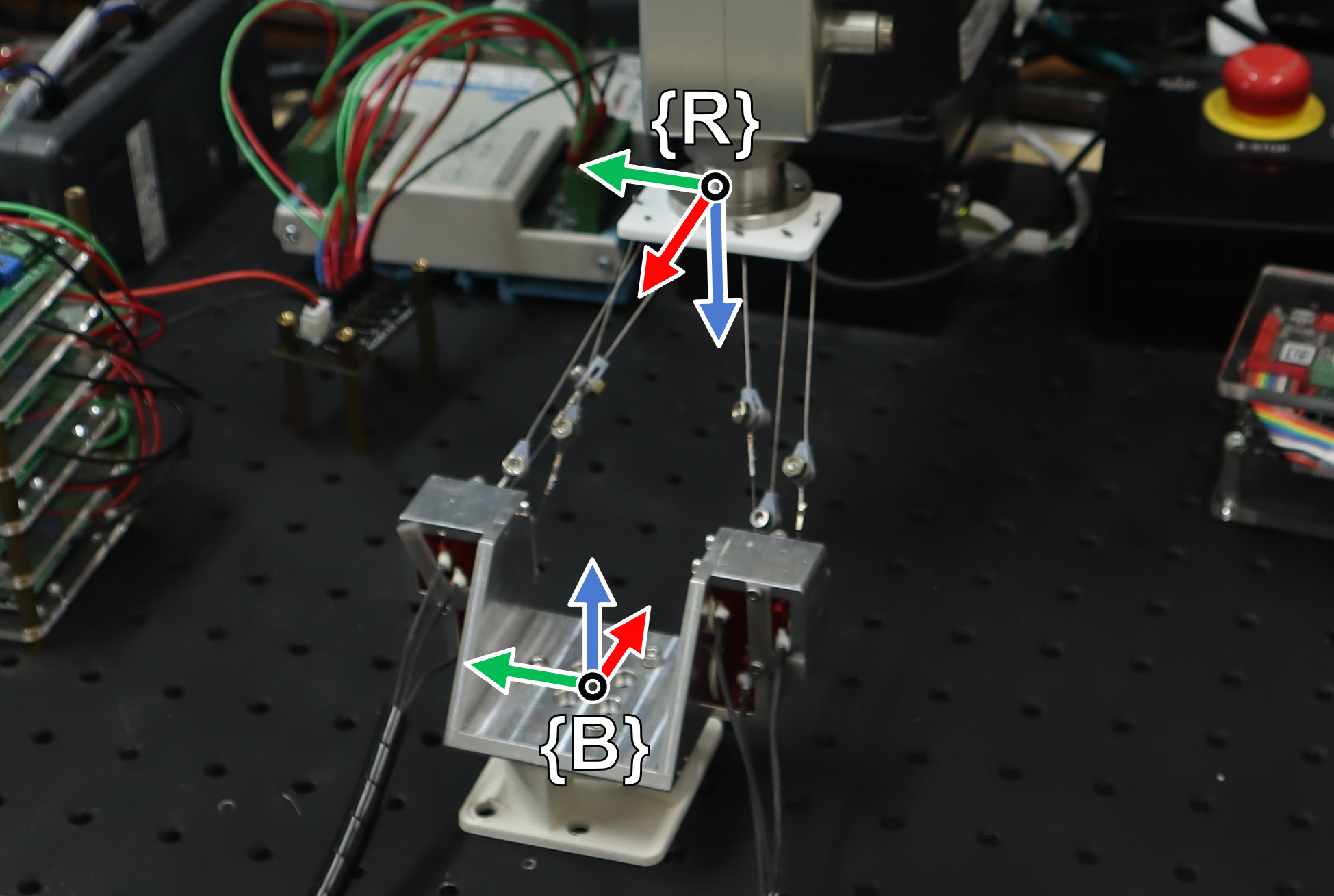}%
  \label{fig: PTM_exp}%
}\quad
\subfloat[Robot-Patient alignment]{%
  \includegraphics[width=5.5cm]{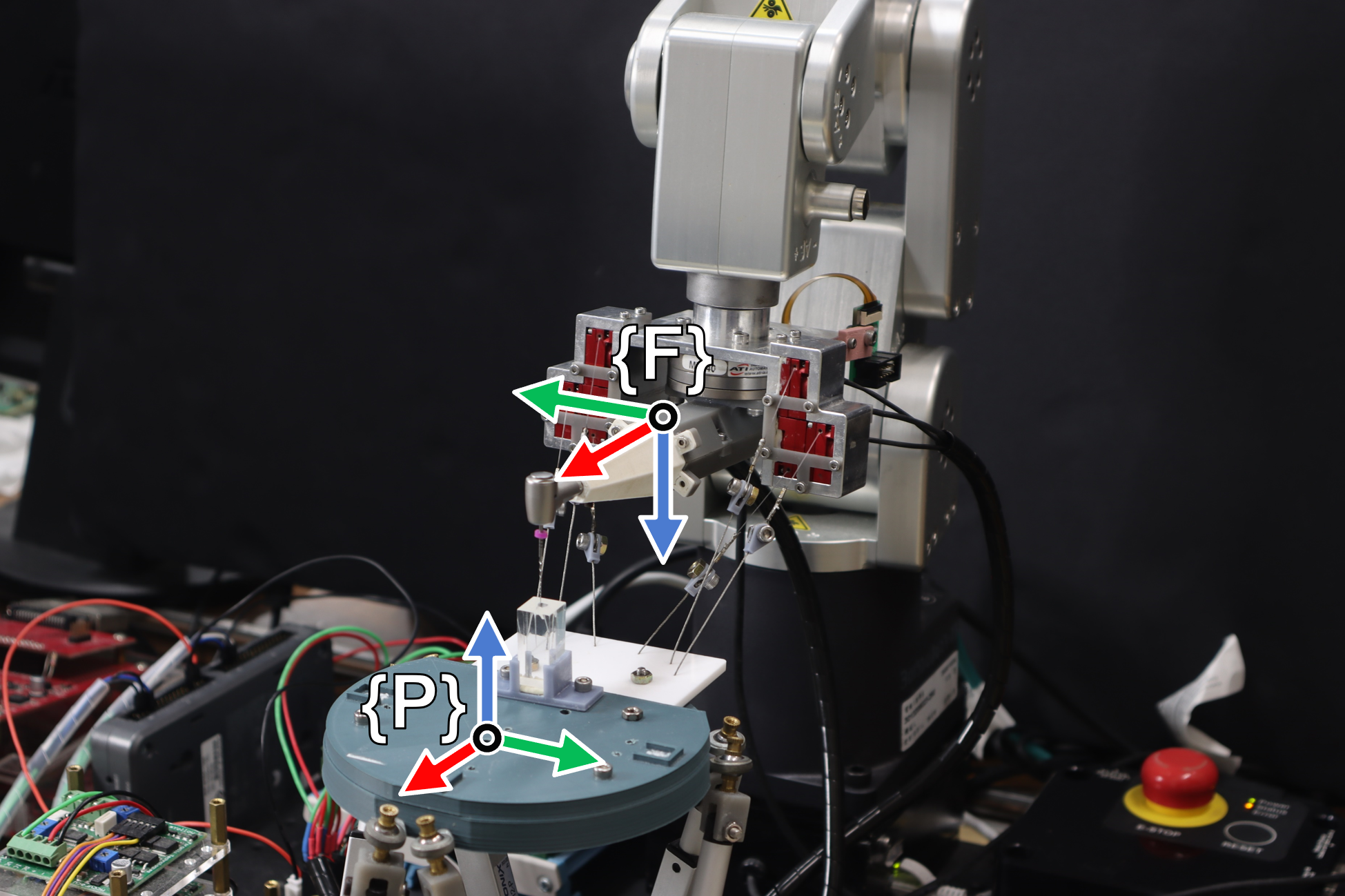}%
  \label{fig: real-time}%
} 
\caption{Experimental setup for the engineering evaluation. (a) For selecting the parameters in the admittance controller and file flexibility compensator, a linear platform was used to create the one-dimensional motion and contact force. (b) For evaluating the accuracy of the PTM, the base of the PTM was fixed on the table. The robot simulated the patient's motion and moved the anchor of the PTM. The measurements obtained from the PTM and the joint encoders of the robotic manipulator were compared. (c) For assessing the performance of robot-patient alignment while executing robot-assisted endodontic treatment, a 6-DoF Stewart platform installed with a root canal model was applied as the patient motion simulator. The DentiBot performed the autonomous procedure and real-time alignment under 6-DoF patient motions.}
\label{fig: exp_setting}
\end{figure*}

The experimental setups for different evaluations are summarized in Fig. \ref{fig: exp_setting}. For the selection of control parameters, our approach involved designing the parameters for the admittance controller, which responds to external contact forces, followed by determining the gain in the file flexibility compensator. Since the endodontic file exhibits symmetry along its x- and y-axes, we utilized a linear platform to generate one-dimensional contact with the endodontic file (Fig. \ref{fig: linear_exp}). Initially, the linear platform was aligned with the file, after which it was moved to exert external force on the file. By employing the admittance controller and file flexibility compensator, the endodontic file effectively tracked the motion of the linear platform. The positions of the linear platform and the endodontic file were recorded using a motion capture system (Impulse X2E, PhaseSpace Inc., San Leandro, CA) with a precision of $0.1$ mm \cite{phasespace}. An active marker was affixed to the top of the dental handpiece, while three additional markers were attached to the linear platform. These three measured positions established a coordinate system $\{P\}$ for the linear platform, where the x-axis represents the primary travel axis. To evaluate the tracking performance, we calculated the tracking error by comparing the distance between the linear platform and the dental handpiece. Our objective was to identify a set of control parameters that minimizes the tracking error.

For the evaluation of the PTM, our objective was to verify its accuracy in measuring both translational and rotational motions. In the experiment, the base of the PTM was securely fixed on the table, while the anchor was attached to the end effector of the 6-DoF robotic manipulator (Fig. \ref{fig: PTM_exp}). The robotic manipulator, with a path accuracy of $0.1$ mm, emulated the patient's motion and controlled the movement of the PTM's anchor. With this setup, the kinematic chain can be expressed as:
\begin{equation}
_{R}^{G}\textbf{T} = \ 
_{B}^{G}\textbf{T}_{A}^{B}\textbf{T}_{R}^{A}\textbf{T}.
\end{equation}
This equation signifies that the PTM's measurements, $_{A}^{B}\textbf{T}$, can be mapped to the motion of the robot arm, $_{R}^{G}\textbf{T}$, by multiplying two precalibrated transformations, $_{R}^{A}\textbf{T}$ and $_{B}^{G}\textbf{T}$. Therefore, in this evaluation, we compared the measurements of $_{R}^{G}\textbf{T}$, one is obtained from the PTM, the other is obtained from the readings of the robot's joint encoders.

To assess the performance of robot-patient alignment during robot-assisted endodontic treatment, we employed a 6-DoF Stewart platform installed with a root canal model as the patient motion simulator (Fig. \ref{fig: real-time}). The same setup was also applied to the preclinical evaluation on root canal models. Here, the patient motion simulator was controlled independently by an NI myRIO-1990 controller. To ensure data integrity, there was no communication between the DentiBot and the patient motion simulator. The DentiBot autonomously executed the procedure and achieved real-time alignment under 6-DoF patient motions. The motion of both the DentiBot and the patient motion simulator was captured using the motion capture system. For coordinate system establishment, seven active markers were mounted on the robotic manipulator and the patient motion simulator, with each device having three markers. An additional marker was placed on the base of the endodontic file to obtain more accurate file position data. The position data were then transformed into the patient's frame $\{P\}$ to calculate the alignment error.

\subsection{Engineering Evaluation} 

\subsubsection{Control Parameter Selection} The design and selection of control parameters for the DentiBot are described in this subsection. There are three controllers in the proposed 6-DoF hybrid position/force control structure (Fig. \ref{fig: overall structure}), namely the PD position controller, the admittance controller, and the file flexibility compensator. First, parameters of the PD controller (Eq. \ref{eq:PD}) for each translational and rotational axis, $k_p$ and $k_d$, were tuned as listed in Tab. \ref{tab: ctrl_param}. 

Regarding the admittance controller (Eq. \ref{eq:adm}), there are three parameters to be designed for each axis: the mass coefficient $m_a$, the damper coefficient $b_a$, and the constant gain $k_a$. The determination of the mass coefficients was based on the estimated mass or inertia of the dental handpiece, obtained from the CAD model. The virtual damper coefficients were designed to ensure that the system settles within a specified time. For this study, a settling time of $0.05$ s was chosen to track the patient's movements. As a result, the time constant of this first-order virtual system was $0.01$ s, requiring a $1:100$ ratio between $m_a$ and $b_a$. The selected parameters for the admittance controller can also be found in Tab. \ref{tab: ctrl_param}.

The constant gain $k_a$ in the admittance controller plays a crucial role in adjusting the system's performance and response speed. A higher value of $k_a$ increases the system's sensitivity to external forces. In the previous version of DentiBot \cite{cheng2022force}, the value of $k_a$ was set to $1.6$ based on the alignment error and force observed during robot-patient alignment experiments, where only the admittance control was utilized. However, in the proposed 6-DoF hybrid position/force control structure, the effects of $k_f$ (the virtual spring constant in the file flexibility compensator (Eq. \ref{eq:fileflexcomp})) are coupled with those of $k_a$. To determine the values of $k_a$ and $k_f$ for the DentiBot, the following experiments were conducted.

\begin{figure}[!t]
\centering
\subfloat[Selection of $k_a$]{%
  \includegraphics[width=4.25cm]{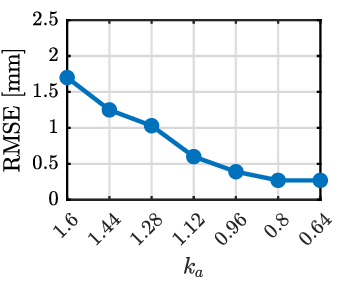}%
  \label{fig: param_sel_ka}%
}
\subfloat[Selection of $k_f$]{%
  \includegraphics[width=4.25cm]{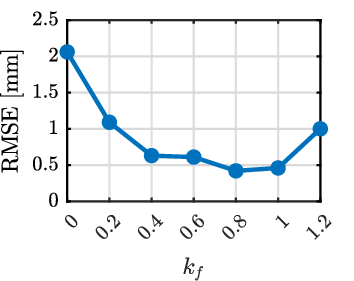}%
  \label{fig: param_sel_kf}%
} 
\caption{Parameter selection of the admittance controller and the file flexibility compensator. (a) The admittance gain $k_{a}$ was selected for the smallest RMS tracking error, i.e., the distance between the robot and the linear moving platform. Note that in this experiment, the insertion depth of the endodontic file was set as $12$ mm such that the file flexibility had a minimal effect on the tracking error. From the experimental results, we selected $k_{a}=0.8$. (b) The virtual spring constant $k_{f}$ was selected for the smallest RMS tracking error when the insertion depth of the endodontic file was set as $0$ mm (only the file tip contacted with the moving platform) and $k_{a}$ equalled $0.8$. Based on the test results, we selected $k_{f}=0.8$ for the following experiments. }
\label{fig: param_sel}
\end{figure}

The selection of the admittance control gain $k_a$ was based on minimizing the tracking error, specifically the distance between the robot and the linear platform (Fig. \ref{fig: linear_exp}) as the linear platform moves. In this experiment, the insertion depth of the endodontic file was fixed at $12$ mm to minimize the influence of file flexibility on the tracking error. The value of $k_a$ was gradually decreased from $1.60$ to $0.64$, while $k_f$ remained fixed at $1$. To ensure the reproducibility of the results, the experiment was repeated five times for each $k_a$ value. The results of the experiment are presented in Fig. \ref{fig: param_sel_ka}. The tracking error consistently decreased until $k_a$ reached $0.8$, which was determined as the optimal value for $k_a$.

\begin{table}[!t]
\caption{Control Parameters of the DentiBot}
\label{tab: ctrl_param}
\centering
\begin{tabular}{c|rr|rrr|r}
\hline\hline
\textbf{Axis} & \multicolumn{1}{c}{$k_p$} & \multicolumn{1}{c|}{$k_d$} & \multicolumn{1}{c}{$m_a$} & \multicolumn{1}{c}{$b_a$} & \multicolumn{1}{c|}{$k_a$} & \multicolumn{1}{c}{$k_f$} \\ \hline
$x$             & $5.0$                        & $0.0015$                    & $0.4$                      & $40$                       & $0.8$                       & $0.8$                      \\
$y$             & $5.0$                        & $0.0015$                    & $0.4$                      & $40$                       & $0.8$                       & $0.8$                      \\
$z$             & $5.0$                        & $0.0015$                    & $0.4$                      & $40$                       & $1.6$                      & $0$                        \\
$\phi$           & $1.5$                      & $0.0005$                    & $0.001157$                 & $0.1157$                   & $1.6$                       & $0$                        \\
$\psi$           & $1.5$                      & $0.0005$                    & $0.001633$                 & $0.1633$                   & $1.6$                       & $0$                        \\
$\theta$         & $1.5$                      & $0.0005$                    & $0.001208$                 & $0.1208$                   & $0$                         & $0$                        \\ \hline\hline
\end{tabular}
\end{table}

Once $k_a$ had been determined, another experiment was conducted to identify the optimal value for the virtual spring constant $k_f$. The range of $k_f$ was varied from $0$ to $1.2$, while $k_a$ was set to $0.8$. The objective was to select the value of $k_f$ that minimizes the tracking error when the insertion depth of the endodontic file was set to $0$ mm, indicating that only the file tip makes contact with the moving platform, resulting in relatively large compensator output. The results of this experiment are illustrated in Fig. \ref{fig: param_sel_kf}. As $k_f$ increased, the tracking error consistently decreased until $k_f$ reached a value of $0.8$. Based on this outcome, $k_f$ was also set to $0.8$ for the DentiBot, as it yielded the optimal position tracking performance.

\begin{figure}[!t]
\centering
\subfloat[Without compensation]{%
  \includegraphics[width=4.25cm]{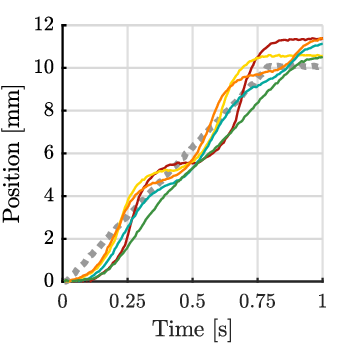}%
  \label{fig: file_no_comp}%
}\quad
\subfloat[With compensation]{%
  \includegraphics[width=4.25cm]{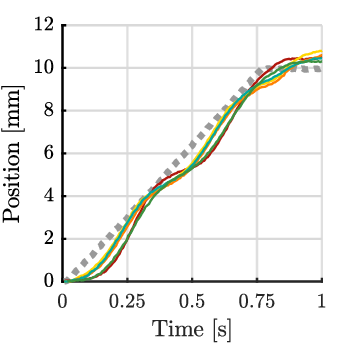}%
  \label{fig: file_comp}%
}
\caption{Evaluation of the file flexibility compensator with various file insertion depths. In this experiment, the linear platform made contact with the endodontic file at different insertion depths and moved along the ramp reference trajectory (\protect\PatientLine). We recorded and compared the motion trajectories of the DentiBot in response to the contact force. File insertion depths: $12$ mm (\protect\InsDeptTwlvLine), $9$ mm (\protect\InsDeptNineLine), $6$ mm (\protect\InsDeptSixLine), $3$ mm (\protect\InsDeptThreeLine), and only contacting the file tip (\protect\InsDeptZeroLine). (a) Before applying the compensator, the robot motion varied with different file insertion depths. In particular, when the insertion depth was set as $12$ mm (\protect\InsDeptTwlvLine), the robot motion was overly driven by the admittance controller, causing a significant oscillation. (b) Upon applying the compensator, the robot's motion trajectories exhibited increased consistency with changes in the file insertion depth. Additionally, the magnitudes of oscillation were reduced.}
\label{fig:file_depth}
\end{figure}

After the parameter selection process, the performance of the system incorporating the file flexibility compensator was evaluated. In this experiment, the linear platform made contact with the endodontic file at various insertion depths and followed a ramp reference trajectory. Specifically, the linear platform moved at a speed of $13.3$ mm/s for a duration of $10$ s before coming to a stop. Five trials were conducted for each case. The experimental results are depicted in Fig. \ref{fig:file_depth}.

In the absence of file flexibility compensation (Fig. \ref{fig: file_no_comp}), severe oscillations were observed when the insertion depth was large, leading to significant errors in the steady state (after $t=0.75$ s). In particular, when the insertion depth was set as $12$ mm, the robot motion was overly driven by the admittance controller. This suggests that the admittance control gain $k_a$ is not suitable for all file insertion depths. Conversely, Fig. \ref{fig: file_comp} demonstrates improved performance after compensating for file flexibility. Despite possible delays in the feedback control loop, the tracking errors were smaller compared to the previous case. The robot's motion trajectories exhibited increased consistency with changes in the file insertion depth, and the magnitudes of oscillation were reduced.

\subsubsection{Accuracy of Patient Tracking Module}
The accuracy of the PTM was evaluated in this subsection. First, the translational measuring error of the PTM was assessed by moving the robotic manipulator along three linear axes respectively, each with a stroke of $15$ mm (see Fig. \ref{fig: PTM_exp}). The average discrepancies between the PTM measurements and the robot's joint encoders are provided in Tab. \ref{tab: PTM_measure}. The errors were all below $0.4$ mm, indicating good accuracy. The error along the z-axis was relatively larger due to the inclusion of errors from both the x- and y-axes during the coordinate transformation between the measurements. Despite this, the translational measuring errors were pretty close to the resolution of the string potentiometers, which is $0.2$ mm. To achieve further improvements, it would be necessary to use a string potentiometer with a higher level of accuracy.

\begin{figure}[!t]
\centering
\subfloat[Translation]{%
  \includegraphics[width=8cm]{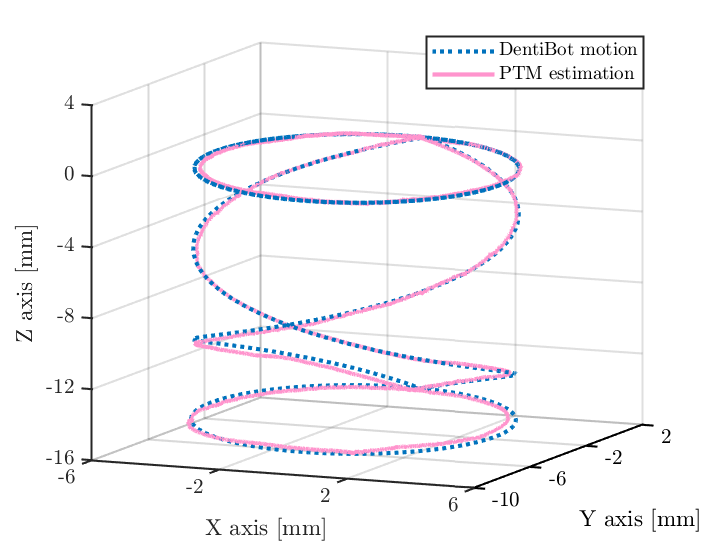}%
  \label{fig: PTM_translation}%
}\quad
\subfloat[Orientation]{%
  \includegraphics[width=8cm]{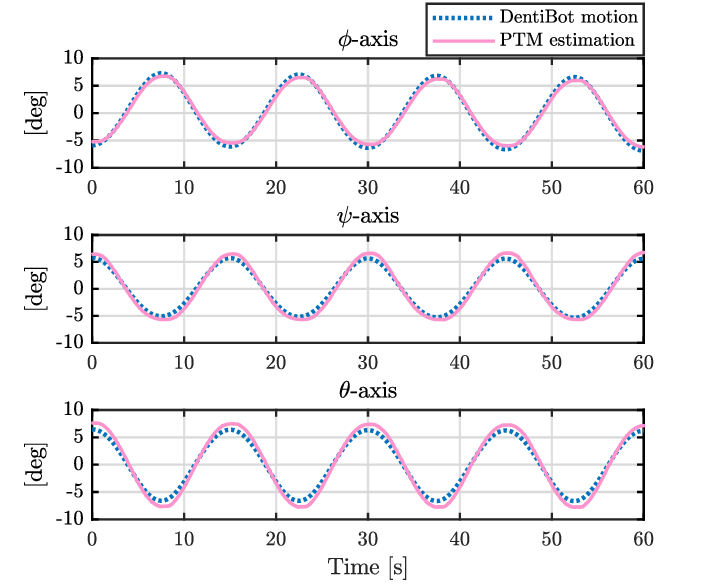}%
  \label{fig: PTM_orientation}%
}
\caption{Evaluation of the measuring accuracy of the PTM. (a) Comparison of position estimations when the robot moved inside the cylindrical workspace. The average error between the measurements from the PTM and that from the joint encoders of the robotic manipulator was $0.40$ mm. (b) Comparison of orientation estimations when the robot was changing its roll, pitch, and yaw angles simultaneously. The average error was $1.15$ deg.}
\label{fig: PTM_evaluation}
\end{figure}

Subsequently, the robotic manipulator was directed to traverse a three-dimensional trajectory spanning the required cylindrical workspace, as shown in Fig. \ref{fig: PTM_translation}. During this trajectory, all linear axes were actuated and measured simultaneously. The PTM demonstrated an average measuring error of $0.4$ mm, with a maximum error of $1.0$ mm observed when the robotic manipulator approached the boundaries of the cylindrical workspace.

\begin{table}[htbp]
\caption{Accuracy evaluation of the PTM}
\centering
\begin{tabular}{cc|cc} 
\hline \hline
\multicolumn{2}{c|}{\textbf{Translational motion}} & \multicolumn{2}{c}{\textbf{Rotational motion}}  \\
\hline
Motion axis & Average error  & Motion axis & Average error \\
             &      [mm]      &            &    [deg]      \\
\hline
x        &$0.27$   &$\phi$       &$0.91$  \\
y        &$0.31$   &$\psi$       &$0.75$  \\
z        &$0.39$   &$\theta$     &$0.74$ \\
x+y+z   &$0.40$  &$\phi$+$\psi$+$\theta$   &$1.15$\\
\hline\hline
\end{tabular}
\label{tab: PTM_measure}
\end{table}

Next, the accuracy of the PTM in measuring rotational motions was evaluated. The robotic manipulator was commanded to track a sinusoidal trajectory, one axis at a time, with an amplitude of $6$ degrees and a period of $15$ seconds. Tab. \ref{tab: PTM_measure} presents the quaternion distances between the PTM measurements and those obtained from the robot's joint encoders. The measuring errors for each axis were all below $1$ degree. Furthermore, an additional experiment was conducted where all three axes simultaneously rotated and followed the sinusoidal trajectory. As shown in Fig. \ref{fig: PTM_orientation}, the average measuring error was found to be less than $1.5$ degrees. These results confirm the effectiveness and performance of the PTM, positioning it as a reliable tool for real-time pose measurements in the context of robot-patient alignment.

\subsubsection{Performance of Robot-Patient Alignment}

This experiment analyzes and compares the performance of different control schemes for robot-patient alignment: (a) Admittance control only, as previously implemented and reported in \cite{cheng2022force}; (b) Admittance control with the file flexibility compensator, without involving the PTM; and (c) the proposed 6-DoF hybrid position/force control. In this experiment, the endodontic file was inserted statically into the acrylic root canal model. 
The patient motion simulator (Fig. \ref{fig: real-time}) moved along a slanted circular trajectory within the required cylindrical workspace, with a radius of $20$ mm and a depth of $20$ mm. Additionally, the roll, pitch, and yaw angles of the patient motion simulator followed a sinusoidal trajectory with an amplitude of $5$ degrees. Various speeds of the 6-DoF patient's motion were tested, with average velocities of $1.5$ mm/s, $2.0$ mm/s, and $2.5$ mm/s, respectively. The alignment error and contact force while using different controllers were recorded and compared.

\begin{figure*}[t]
\centering
  \includegraphics[trim={1.5cm 0.8cm 1.5cm 0.2cm},clip,width=18cm]{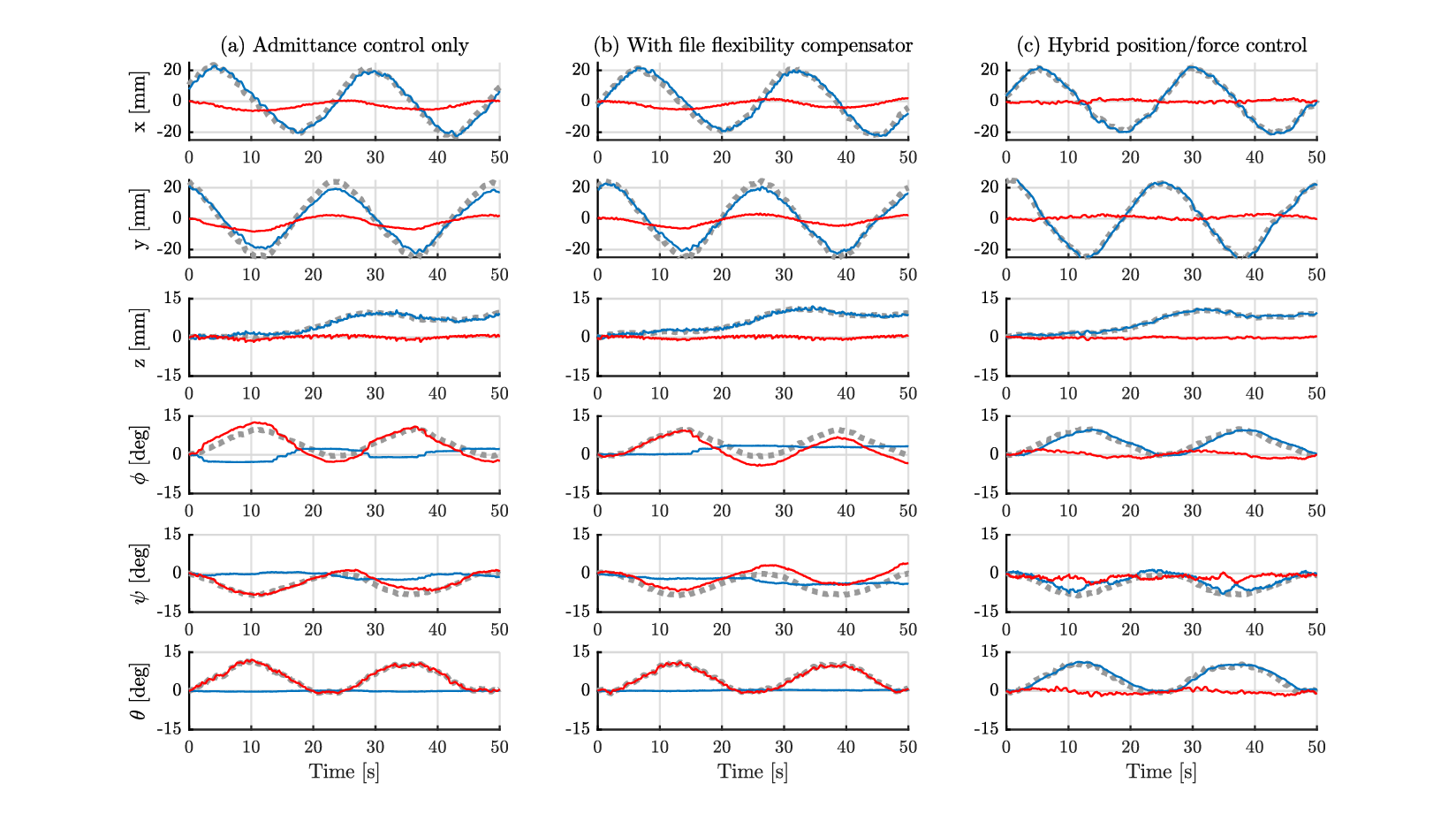}
\caption{Comparison of real-time robot-patient alignment under different control methods: (a) Admittance control only; (b) Admittance control with the file flexibility compensator; and (c) the proposed 6-DoF hybrid position/force control. The intraoperative patient's motions (\protect\PatientLine) were simulated by the 6-DoF patient motion simulator. As indicated by the alignment errors (\protect\ErrLine), the DentiBot (\protect\DentiBotLine) was unable to adapt to the patient's attitude changes using only the admittance controller, and there were minor delays in the translational DoFs. While the file flexibility compensator could decrease the force and torque exerted on the endodontic file, it did not enhance the alignment accuracy. However, the proposed hybrid position/force controller addressed alignment errors in both translational and rotational DoFs.}
\label{fig: real-time alignment}
\vspace{-3mm}
\end{figure*}

The alignment errors between the robot and the patient using different controllers are documented in Tab. \ref{tab:alignment-position}. Fig. \ref{fig: real-time alignment} illustrates the trajectories of both the patient motion simulator and the DentiBot, with an average speed of $2.0$ mm/s. Comparing the alignment errors under the first two control schemes, namely (a) Admittance control only and (b) Admittance control with file flexibility compensator, it can be observed that the latter exhibited a slight reduction in both translational and rotational DoFs. The speed of the motion did not significantly affect the experimental results. However, Fig. \ref{fig: real-time alignment} indicates that both controllers exhibited a minor phase delay when tracking translational motions. Additionally, neither controller adequately addressed the rotational motions due to the conical shape of the root canal. Notably, the DentiBot failed to align with the patient in the $\theta$-axis because the admittance controller was intentionally disabled, with the gain $k_a$ for this rotational DoF set to zero. This was done to prevent the torque generated by file rotation from interfering with the robot-patient alignment.

\begin{table}
\caption{Root-Mean-Square Tracking Errors During Robot-Patient Alignment Experiment}
\label{tab:alignment-position}
\centering
\begin{tabular}{ccccccc}
\hline\hline
\multicolumn{7}{c}{\textbf{(a) Admittance Control Only}}                                                                                        \\ \hline
\multicolumn{1}{c|}{Speed}      & $e_x$         & $e_y$         & \multicolumn{1}{c|}{$e_z$}         & $e_\phi$      & $e_\psi$      & $e_\theta$    \\
\multicolumn{1}{c|}{{[}mm/s{]}} & \multicolumn{3}{c|}{{[}mm{]}}                                      & \multicolumn{3}{c}{{[}deg{]}}                 \\ \hline
\multicolumn{1}{c|}{2.5}        & 3.03          & 4.43          & \multicolumn{1}{c|}{0.79}          & 6.89          & 4.56          & 3.86          \\
\multicolumn{1}{c|}{2.0}        & 3.48          & 4.18          & \multicolumn{1}{c|}{0.54}          & 6.49          & 4.72          & 3.94          \\
\multicolumn{1}{c|}{1.5}        & 3.71          & 2.97          & \multicolumn{1}{c|}{0.50}          & 4.82          & 5.22          & 3.95          \\ \hline\hline
\multicolumn{7}{c}{\textbf{(b) Adm. Ctrl with File Flex. Compensator}}                                                                          \\ \hline
\multicolumn{1}{c|}{Speed}      & $e_x$         & $e_y$         & \multicolumn{1}{c|}{$e_z$}         & $e_\phi$      & $e_\psi$      & $e_\theta$    \\
\multicolumn{1}{c|}{{[}mm/s{]}} & \multicolumn{3}{c|}{{[}mm{]}}                                      & \multicolumn{3}{c}{{[}deg{]}}                 \\ \hline
\multicolumn{1}{c|}{2.5}        & 3.13          & 3.42          & \multicolumn{1}{c|}{0.62}          & 5.25          & 4.23          & 4.11          \\
\multicolumn{1}{c|}{2.0}        & 3.28          & 2.52          & \multicolumn{1}{c|}{0.47}          & 3.98          & 4.36          & 4.12          \\
\multicolumn{1}{c|}{1.5}        & 3.61          & 2.59          & \multicolumn{1}{c|}{0.52}          & 4.04          & 5.29          & 4.05          \\ \hline\hline
\multicolumn{7}{c}{\textbf{(c) Hybrid Position/Force Control}}                                                                                  \\ \hline
\multicolumn{1}{c|}{Speed}      & $e_x$         & $e_y$         & \multicolumn{1}{c|}{$e_z$}         & $e_\phi$      & $e_\psi$      & $e_\theta$    \\
\multicolumn{1}{c|}{{[}mm/s{]}} & \multicolumn{3}{c|}{{[}mm{]}}                                      & \multicolumn{3}{c}{{[}deg{]}}                 \\ \hline
\multicolumn{1}{c|}{2.5}        & \textbf{1.50} & \textbf{1.27} & \multicolumn{1}{c|}{\textbf{0.30}} & \textbf{1.87} & \textbf{2.05} & \textbf{1.31} \\
\multicolumn{1}{c|}{2.0}        & \textbf{1.45} & \textbf{1.36} & \multicolumn{1}{c|}{\textbf{0.32}} & \textbf{2.05} & \textbf{2.06} & \textbf{1.00} \\
\multicolumn{1}{c|}{1.5}        & \textbf{0.96} & \textbf{1.26} & \multicolumn{1}{c|}{\textbf{0.24}} & \textbf{1.80} & \textbf{1.25} & \textbf{1.08} \\ \hline\hline
\end{tabular}
\end{table}

\begin{table}
\caption{ Average Forces/Torques Applied to the Endodontic File During Robot-Patient Alignment Experiment}
\centering
\begin{tabular}{ccccccc}
\hline\hline
\multicolumn{7}{c}{\textbf{(a) Admittance Control Only}}                                                                \\ \hline
\multicolumn{1}{c|}{Speed}      & $f_x$         & $f_y$ & \multicolumn{1}{c|}{$f_z$} & $\tau_x$         & $\tau_y$    & $\tau_z$   \\
\multicolumn{1}{c|}{{[}mm/s{]}} & \multicolumn{3}{c|}{{[}N{]}}                       & \multicolumn{3}{c}{{[}mN $\cdot$ m{]}} \\ \hline
\multicolumn{1}{c|}{2.5}        & {0.37} & 0.37  & \multicolumn{1}{c|}{\textbf{0.43}}  & {4.23}    & 3.80     & \textbf{1.36}    \\
\multicolumn{1}{c|}{2.0}        & {0.30} & 0.29  & \multicolumn{1}{c|}{\textbf{0.43}}  & {3.41}    & 3.07     & \textbf{1.28}    \\
\multicolumn{1}{c|}{1.5}        & {0.24} & 0.23  & \multicolumn{1}{c|}{0.44}  & {2.68}    & 2.36     & 2.16    \\ \hline\hline
\multicolumn{7}{c}{\textbf{(b) Adm. Ctrl with File Flex. Compensator}}                                   \\ \hline
\multicolumn{1}{c|}{Speed}      & $f_x$         & $f_y$ & \multicolumn{1}{c|}{$f_z$} & $\tau_x$         & $\tau_y$    & $\tau_z$   \\
\multicolumn{1}{c|}{{[}mm/s{]}} & \multicolumn{3}{c|}{{[}N{]}}                       & \multicolumn{3}{c}{{[}mN $\cdot$ m{]}} \\ \hline
\multicolumn{1}{c|}{2.5}        & \textbf{0.32} & \textbf{0.27}  & \multicolumn{1}{c|}{\textbf{0.43}}  & \textbf{3.07}    & \textbf{3.06}     & 1.57    \\
\multicolumn{1}{c|}{2.0}        & \textbf{0.26} & \textbf{0.22}  & \multicolumn{1}{c|}{\textbf{0.43}}  & \textbf{2.42}    & \textbf{2.49}     & 1.41    \\
\multicolumn{1}{c|}{1.5}        & \textbf{0.22} & \textbf{0.17}  & \multicolumn{1}{c|}{\textbf{0.43}}  & \textbf{2.04}    & \textbf{2.15}     & \textbf{1.08}    \\ \hline\hline
\multicolumn{7}{c}{\textbf{(c) Hybrid Position/Force Control}}                                                          \\ \hline
\multicolumn{1}{c|}{Speed}      & $f_x$         & $f_y$ & \multicolumn{1}{c|}{$f_z$} & $\tau_x$         & $\tau_y$    & $\tau_z$   \\
\multicolumn{1}{c|}{{[}mm/s{]}} & \multicolumn{3}{c|}{{[}N{]}}                       & \multicolumn{3}{c}{{[}mN $\cdot$ m{]}} \\ \hline
\multicolumn{1}{c|}{2.5}        & {0.47} & 0.47  & \multicolumn{1}{c|}{0.50}  & {5.60}    & 5.06     & 1.47    \\
\multicolumn{1}{c|}{2.0}        & {0.40} & 0.35  & \multicolumn{1}{c|}{0.49}  & {4.28}    & 4.46     & 1.32    \\
\multicolumn{1}{c|}{1.5}        & {0.34} & 0.28  & \multicolumn{1}{c|}{0.48}  & {3.54}    & 3.79     & 1.23    \\ \hline\hline
\end{tabular}
\label{tab:alignment-force }
\end{table}

Next, the alignment errors when using (b) Admittance control with file flexibility compensator and (c) 6-DoF hybrid position/force controller are compared. The integration of the PTM into the DentiBot resulted in significant improvements in all axes. The phase delay in translational DoFs was eliminated, leading to alignment errors below $1.50$ mm. Furthermore, the DentiBot successfully tracked the patient's rotational motions with average errors of approximately $1$-$2$ degrees, which was about three times smaller than when using the other two controllers. The system specification (S3) regarding the error tolerance for patient motion tracking was met. These results clearly demonstrate the substantial enhancement in real-time alignment performance of the DentiBot achieved by incorporating the PTM and employing the hybrid position/force control approach.

The average force/torque applied to the endodontic file during the alignment experiments is summarized in Tab. \ref{tab:alignment-force }. It is observed that both the force and torque increased in correlation with the velocity of the patient's movement, which aligned with expectations. When comparing the different controllers, it is evident that (b) Admittance control with file flexibility compensator resulted in the lowest contact force. This indicates that the combination of the file flexibility compensator has a better ability to reduce the contact force. With the addition of the PTM, there was an increase in both contact force and torque. This increase can be attributed to the fact that the PTM was not aware of the curvature of the root canal, which may lead to unexpected impacts during patient tracking. However, despite this increase, the contact force and torque still met the requirements specified in the system specification (S2).

\subsection{Pre-clinical Evaluation}

This subsection presents an evaluation of the performance of autonomous cleaning and shaping when using the DentiBot in RCT. The surgical workflow depicted in Fig. \ref{fig:finite_machine} was employed. During the pre-clinical evaluation, the patient motion simulator generated 6-DoF motions as the one applied in the engineering evaluation (Fig. \ref{fig: real-time}). To ensure precise file insertion while maintaining the relative pose between the robot and the patient, the proposed 6-DoF hybrid position/force control method was activated. 

\begin{figure}[!b]
\centering
\subfloat[Acrylic model]{%
  \includegraphics[width=4cm]{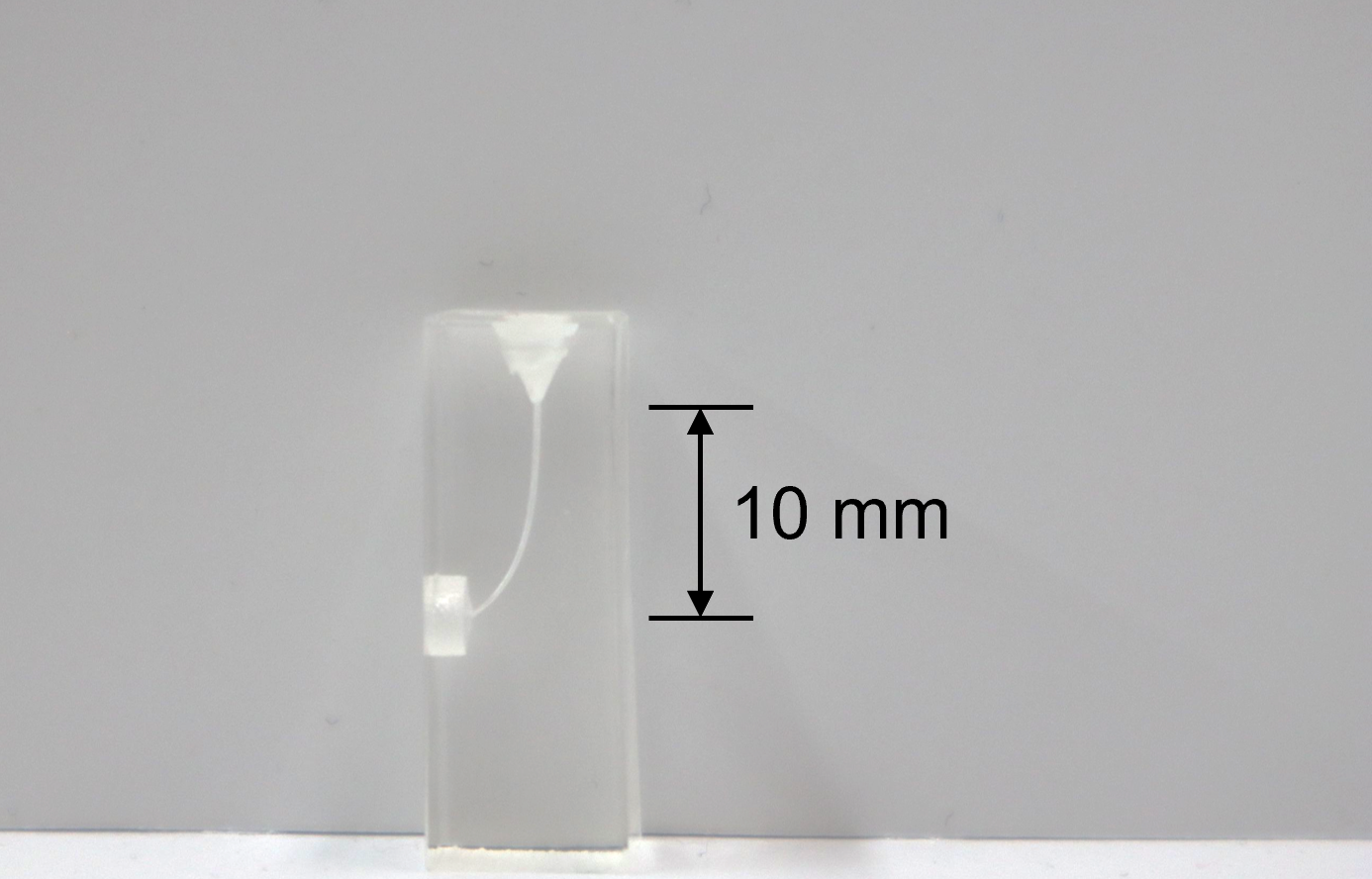}%
  \label{fig: arcrylic_root_module}%
}\quad
\subfloat[AA temp resin teeth]{%
  \includegraphics[width=4cm]{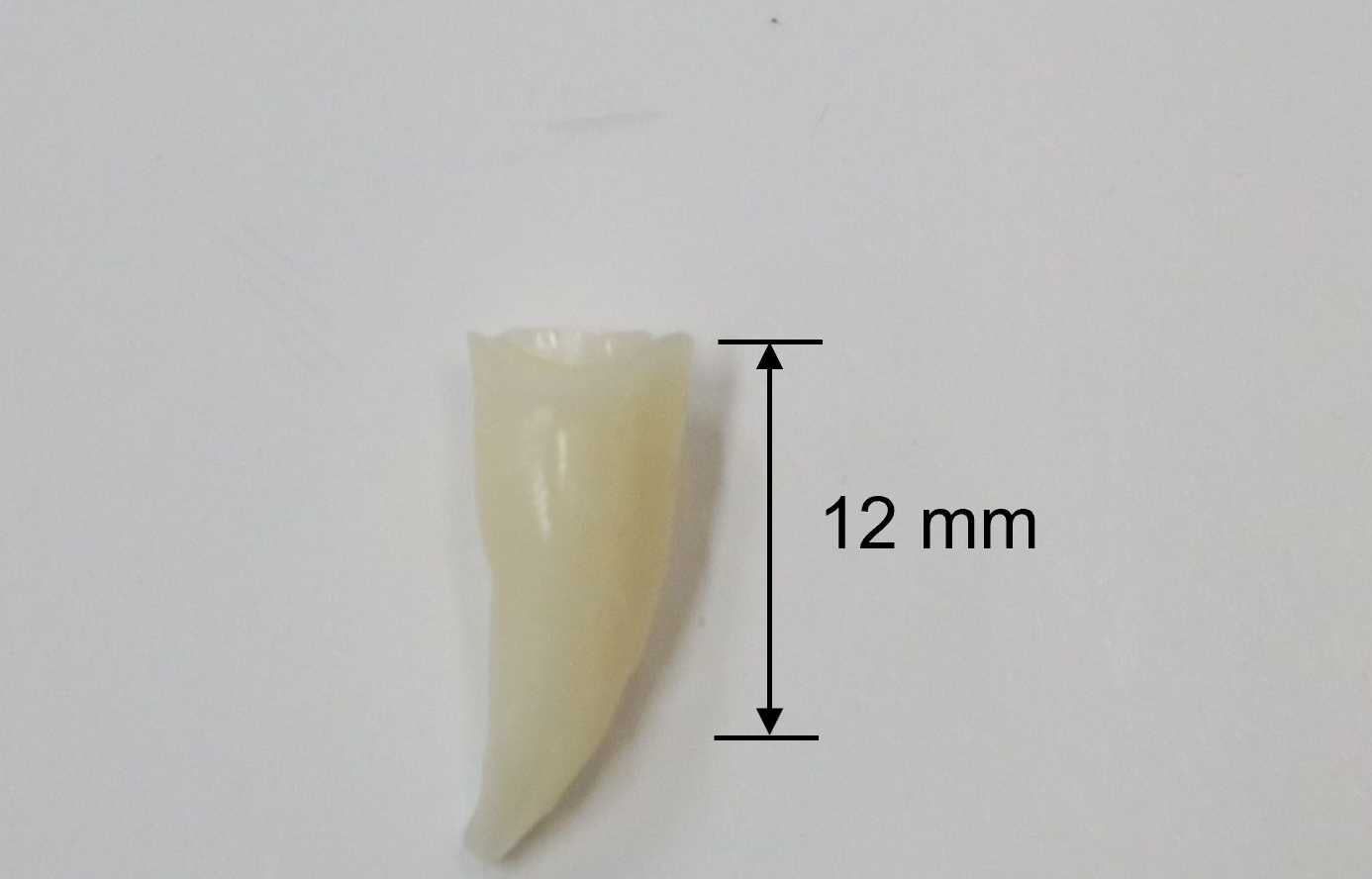}%
  \label{fig: 3d_printed_root_module}%
} 
\caption{Root canal models applied in the pre-clinical experiment: (a) Acrylic model and (b) 3D-printed AA temp resin teeth. Both models are commonly used in clinical practice. }
\label{fig: root_module}
\end{figure}

The evaluation involved testing two types of root canal models, as depicted in Fig. \ref{fig: root_module}. Model A was an acrylic root canal model, which is commonly used for dental training. Model B was AA temp resin teeth, which were created by 3D printing denture materials based on CT scans of human root canals. For each type, five models with subtle difference are tested. All the models underwent micro-CT scanning before and after the robot-assisted procedure for further quantitative analysis. Prior to the autonomous procedure, the dentist selected the appropriate endodontic file type. In the case of the acrylic model, the procedure involved using three files (SX, S1, and S2) with increasing diameters. For the resin teeth, three additional files (F1, F2, and F3) were used to further enlarge the root canal cavity.

The results of the pre-clinical evaluation are summarized in Tab. \ref{tab: pre-clinical results}. Overall, all models successfully underwent the complete procedure without any instances of file fracturing or ledging. The acrylic models exhibited a larger enlargement ratio, which can be attributed to their narrower root canal pathway. Importantly, there was no disparity in the root canal length after treatment, indicating the efficacy of intraoperative monitoring of file insertion depth. However, in some models, a slight reduction in root canal length was observed following the autonomous procedure, likely due to the presence of debris deposited near the apex.

\begin{figure}[h]
\centering
  \includegraphics[width=6.5cm]{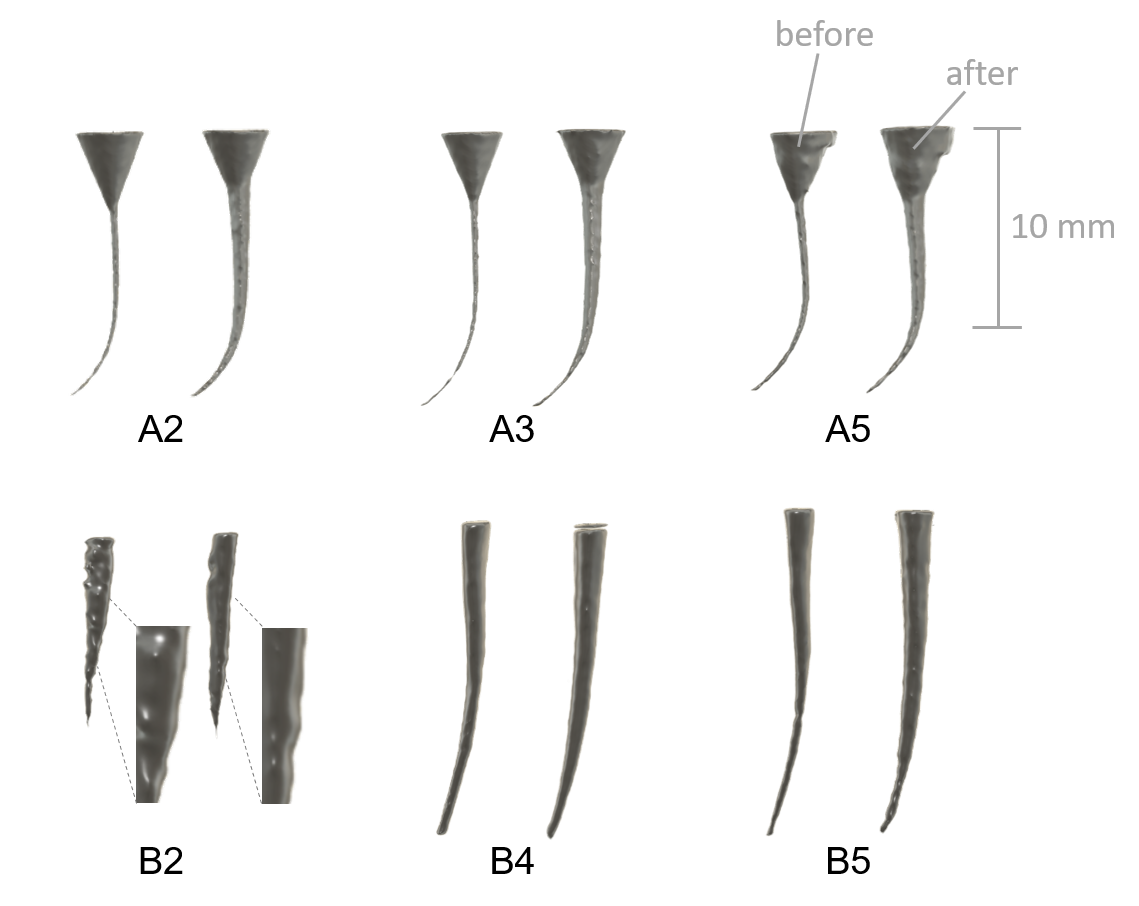}%
\caption{CT scans of the representative root canal models tested in the pre-clinical evaluation. Model A denotes acrylic models and Model B denotes AA temp resin models. The root canals were enlarged and well-shaped after the robot-assisted endodontic treatment. In particular, the wall of the root canal in Model B2 was shown smoother after autonomous root canal cleaning and shaping.}
\label{fig: preclinical_result}
\end{figure}

Fig. \ref{fig: preclinical_result} illustrates the CT scan results of the representative root canal models examined in the pre-clinical evaluation. Comparing the acrylic models to the AA temp resin models, it is evident that the former exhibited a greater degree of curvature. Nevertheless, the DentiBot successfully completed the cleaning and shaping process autonomously, reaching the apex with the endodontic file. As illustrated in Fig. \ref{fig: preclinical_result}, the root canals were effectively enlarged and well-shaped following the robot-assisted procedure. Notably, Model B2 displayed a smoother wall of the root canal after treatment. These results have proven the DentiBot's capability to autonomously perform the cleaning and shaping step in RCT.

\begin{table}[htbp]
\caption{Pre-clinical evaluation results}
\label{tab: pre-clinical results}
\centering
\begin{tabular}{ccccc}
\hline\hline
\textbf{\begin{tabular}[c]{@{}c@{}}Model\\ ID\end{tabular}} & \textbf{\begin{tabular}[c]{@{}c@{}}Root Canal \\ Volume\\ before/after\\ {[}mm$^3${]}\end{tabular}} & \textbf{\begin{tabular}[c]{@{}c@{}}Enlarge\\ Ratio\end{tabular}} & \textbf{\begin{tabular}[c]{@{}c@{}}Root Canal \\ Length\\ before/after\\ {[}mm{]}\end{tabular}} & \textbf{\begin{tabular}[c]{@{}c@{}}Sucess?\\ {[}Y/N{]}\end{tabular}} \\ \hline
A1                                                          &     1.73 / 3.28                                                                                        & 1.90                                                            & 10.76 / 10.80                                                                                     & Y                                                                    \\
A2                                                          &    1.17 / 3.73                                                                                        & 3.16                                                             & 11.88 / 11.93                                                                                     & Y                                                              \\
A3                                                          &    1.28 / 3.51                                                                                        & 2.73                                                             & 11.39 / 11.57                                                                                     & Y                                                                    \\
A4                                                          &   1.35 / 3.87                                                                                         & 2.87                                                             & 12.02 / 11.93                                                                                     & Y                                                                    \\
A5                                                          &   1.46 / 3.92                                                                                        & 2.67                                                             & 12.21 / 12.25                                                                                     & Y                                                                    \\ \hline
B1                                                          &   4.10 / 4.92                                                                                                 & 1.19                                                             & 8.03 / 8.15                                                                              & Y                                                                    \\
B2                                                          &   4.04 / 5.07                                                                                             & 1.25                                                             & 9.08 / 9.14                                                                                       & Y                                                                    \\
B3                                                          &   17.4 / 18.13                                                                                          & 1.04                                                             & 13.74 / 13.33                                                                                     & Y                                                                    \\
B4                                                          &   9.29 / 10.21                                                                                           & 1.09                                                             & 12.84 / 12.96                                                                                     & Y                                                                    \\
B5                                                          &  9.17 / 10.52                                                                                            & 1.14                                                             & 14.63 / 14.14                                                                                     & Y                                                                    \\ \hline\hline
\end{tabular}
\end{table}

\section{Conclusions}
\label{sec:conclu}


This paper presents the design and evaluation of the DentiBot system, which is the first medical robot designed for assisting endodontic treatment. The clinical needs for an adequate workspace, real-time force-guidance, and accurate patient motion tracking are effectively addressed through the integration of a 6-DoF robotic manipulator, a 6-axis force/torque sensor, and a string-based patient tracking module (PTM), respectively. In particular, the proposed file flexibility compensator addresses the bending of endodontic file during the cleaning and shaping procedure. The novel PTM enables accurate estimation of the patient's 6-DoF pose, offering a compact footprint that ensures minimal interference with other surgical equipment. Furthermore, the adoption of a hybrid position/force control scheme facilitates autonomous adjustment of the endodontic file insertion path while maintaining the relative pose between the robot and the patient.

Through extensive experimentation, it has been demonstrated that the hybrid control approach achieves superior alignment between the robot and the patient compared to force control alone, specifically in the context of robot-assisted endodontic treatment. This improvement can be attributed to the unique conical shape of root canals, which differs from conventional dental implant surgery. The pre-clinical evaluation conducted on acrylic models and resin teeth has provided compelling evidence that the DentiBot system can autonomously and effectively clean and shape root canals, ensuring safety by avoiding file fracturing or ledging. 

As part of future work, the system will be enhanced by integrating automated file selection and an apex locator. Additionally, efforts will be dedicated to automating other essential surgical steps in RCT, including access opening and obturation. These advancements will further augment the capabilities of the DentiBot system and contribute to the advancement of endodontic treatment techniques.


\section*{Acknowledgement}
The authors would like to express their gratitude to Yi-Chan Li, Zi-Hao Huang, Mu-Ting Mao, Shao-Lun Cheng, Chia-Ying Lin, and Dr. Chen-Yu Chan for their invaluable technical support in this project.

\bibliographystyle{IEEEtran}
\bibliography{ref.bib}

\begin{thebibliography}{10}
\providecommand{\url}[1]{#1}
\csname url@samestyle\endcsname
\providecommand{\newblock}{\relax}
\providecommand{\bibinfo}[2]{#2}
\providecommand{\BIBentrySTDinterwordspacing}{\spaceskip=0pt\relax}
\providecommand{\BIBentryALTinterwordstretchfactor}{4}
\providecommand{\BIBentryALTinterwordspacing}{\spaceskip=\fontdimen2\font plus
\BIBentryALTinterwordstretchfactor\fontdimen3\font minus \fontdimen4\font\relax}
\providecommand{\BIBforeignlanguage}[2]{{%
\expandafter\ifx\csname l@#1\endcsname\relax
\typeout{** WARNING: IEEEtran.bst: No hyphenation pattern has been}%
\typeout{** loaded for the language `#1'. Using the pattern for}%
\typeout{** the default language instead.}%
\else
\language=\csname l@#1\endcsname
\fi
#2}}
\providecommand{\BIBdecl}{\relax}
\BIBdecl

\bibitem{gerhardus2003robot}
D.~Gerhardus, ``Robot-assisted surgery: the future is here,'' \emph{Journal of Healthcare Management}, vol.~48, no.~4, p. 242, 2003.

\bibitem{mattos2016microsurgery}
L.~S. Mattos, D.~G. Caldwell, G.~Peretti, F.~Mora, L.~Guastini, and R.~Cingolani, ``Microsurgery robots: addressing the needs of high-precision surgical interventions,'' \emph{Swiss medical weekly}, vol. 146, no. 4344, pp. w14\,375--w14\,375, 2016.

\bibitem{taylor1999steady}
R.~Taylor, P.~Jensen, L.~Whitcomb, A.~Barnes, R.~Kumar, D.~Stoianovici, P.~Gupta, Z.~Wang, E.~Dejuan, and L.~Kavoussi, ``A steady-hand robotic system for microsurgical augmentation,'' \emph{The International Journal of Robotics Research}, vol.~18, no.~12, pp. 1201--1210, 1999.

\bibitem{de2016release}
M.~D. de~Smet, J.~M. Stassen, T.~C. Meenink, T.~Janssens, V.~Vanheukelom, G.~J. Naus, M.~J. Beelen, and B.~Jonckx, ``Release of experimental retinal vein occlusions by direct intraluminal injection of ocriplasmin,'' \emph{British Journal of Ophthalmology}, vol. 100, no.~12, pp. 1742--1746, 2016.

\bibitem{wilson2018intraocular}
J.~T. Wilson, M.~J. Gerber, S.~W. Prince, C.-W. Chen, S.~D. Schwartz, J.-P. Hubschman, and T.-C. Tsao, ``Intraocular robotic interventional surgical system ({IRISS}): Mechanical design, evaluation, and master--slave manipulation,'' \emph{The International Journal of Medical Robotics and Computer Assisted Surgery}, vol.~14, no.~1, p. e1842, 2018.

\bibitem{edwards2018first}
T.~Edwards, K.~Xue, H.~Meenink, M.~Beelen, G.~Naus, M.~Simunovic, M.~Latasiewicz, A.~Farmery, M.~De~Smet, and R.~MacLaren, ``First-in-human study of the safety and viability of intraocular robotic surgery,'' \emph{Nature biomedical engineering}, vol.~2, no.~9, pp. 649--656, 2018.

\bibitem{ahmad2021dental}
P.~Ahmad, M.~K. Alam, A.~Aldajani, A.~Alahmari, A.~Alanazi, M.~Stoddart, and M.~G. Sghaireen, ``Dental robotics: a disruptive technology,'' \emph{Sensors}, vol.~21, no.~10, p. 3308, 2021.

\bibitem{liu2023robotics}
L.~Liu, M.~Watanabe, and T.~Ichikawa, ``Robotics in dentistry: A narrative review,'' \emph{Dentistry Journal}, vol.~11, no.~3, p.~62, 2023.

\bibitem{van2021robot}
T.~C. van Riet, K.~T. C.~J. Sem, J.-P.~T. Ho, R.~Spijker, J.~Kober, and J.~de~Lange, ``Robot technology in dentistry, part two of a systematic review: an overview of initiatives,'' \emph{Dental Materials}, vol.~37, no.~8, pp. 1227--1236, 2021.

\bibitem{bolding2021accuracy}
S.~L. Bolding and U.~N. Reebye, ``Accuracy of haptic robotic guidance of dental implant surgery for completely edentulous arches,'' \emph{The Journal of prosthetic dentistry}, vol. 128, no.~4, pp. 639--647, 2022.

\bibitem{sun2014automated}
X.~Sun, Y.~Yoon, J.~Li, and F.~D. McKenzie, ``Automated image-guided surgery for common and complex dental implants,'' \emph{Journal of medical engineering \& technology}, vol.~38, no.~5, pp. 251--259, 2014.

\bibitem{li2019compact}
J.~Li, Z.~Shen, W.~Y.~T. Xu, W.~Y.~H. Lam, R.~T.~C. Hsung, E.~H.~N. Pow, K.~Kosuge, and Z.~Wang, ``A compact dental robotic system using soft bracing technique,'' \emph{IEEE Robotics and Automation Letters}, vol.~4, no.~2, pp. 1271--1278, 2019.

\bibitem{feng2022image}
Y.~Feng, J.~Fan, B.~Tao, S.~Wang, J.~Mo, Y.~Wu, Q.~Liang, and X.~Chen, ``An image-guided hybrid robot system for dental implant surgery,'' \emph{International Journal of Computer Assisted Radiology and Surgery}, vol.~17, no.~1, pp. 15--26, 2022.

\bibitem{wang2022automation}
H.~Wang, ``Automation in dental and eye surgery,'' Ph.D. dissertation, University of California, Los Angeles, 2022.

\bibitem{wang2014preliminary}
D.~Wang, L.~Wang, Y.~Zhang, P.~Lv, Y.~Sun, and J.~Xiao, ``Preliminary study on a miniature laser manipulation robotic device for tooth crown preparation,'' \emph{The International Journal of Medical Robotics and Computer Assisted Surgery}, vol.~10, no.~4, pp. 482--494, 2014.

\bibitem{yuan2016automatic}
F.~Yuan, Y.~Wang, Y.~Zhang, Y.~Sun, D.~Wang, and P.~Lyu, ``An automatic tooth preparation technique: A preliminary study,'' \emph{Scientific Reports}, vol.~6, no.~1, p. 25281, 2016.

\bibitem{yan2022optics}
B.~Yan, W.~Zhang, L.~Cai, L.~Zheng, K.~Bao, Y.~Rao, L.~Yang, W.~Ye, P.~Guan, W.~Yang \emph{et~al.}, ``Optics-guided robotic system for dental implant surgery,'' \emph{Chinese Journal of Mechanical Engineering}, vol.~35, no.~1, p.~55, 2022.

\bibitem{kim2009study}
G.~Kim, H.~Seo, S.~Im, D.~Kang, and S.~Jeong, ``A study on simulator of human-robot cooperative manipulator for dental implant surgery,'' in \emph{2009 IEEE international symposium on industrial electronics}.\hskip 1em plus 0.5em minus 0.4em\relax IEEE, 2009, pp. 2159--2164.

\bibitem{iijima2020development}
T.~Iijima, T.~Matsunaga, T.~Shimono, K.~Ohnishi, S.~Usuda, and H.~Kawana, ``Development of a multi dof haptic robot for dentistry and oral surgery,'' in \emph{2020 IEEE/SICE International Symposium on System Integration (SII)}.\hskip 1em plus 0.5em minus 0.4em\relax IEEE, 2020, pp. 52--57.

\bibitem{zheng2007computer}
G.~Zheng, L.~Gu, X.~Li, and J.~Zhang, ``Computer-assisted preoperative planning and surgical navigation system in dental implantology,'' in \emph{2007 6th international special topic conference on information technology applications in biomedicine}.\hskip 1em plus 0.5em minus 0.4em\relax IEEE, 2007, pp. 139--142.

\bibitem{sin2023development}
M.~Sin, J.~H. Cho, H.~Lee, K.~Kim, H.~S. Woo, and J.-M. Park, ``Development of a real-time 6-dof motion-tracking system for robotic computer-assisted implant surgery,'' \emph{Sensors}, vol.~23, no.~5, p. 2450, 2023.

\bibitem{daon2014system}
E.~Daon and M.~G. Beckett, ``System and method for determining the three-dimensional location and orientation of identification markers,'' Dec.~9 2014, {US} Patent 8,908,918.

\bibitem{Dong2007WIPAS}
J.~Dong, S.~Hong, and G.~Hesselgren, ``{WIP}: a study on the development of endodontic micro robot,'' in \emph{Proceedings of the 2006 IJME-INTERTECH Conference}, 2006, pp. 104--110.

\bibitem{cheng2022force}
H.-F. Cheng, Y.-C. Li, Y.-C. Ho, and C.-W. Chen, ``Force-guided alignment and file feedrate control for robot-assisted endodontic treatment,'' in \emph{2022 IEEE/RSJ International Conference on Intelligent Robots and Systems (IROS)}.\hskip 1em plus 0.5em minus 0.4em\relax IEEE, 2022, pp. 1841--1847.

\bibitem{hulsmann2005mechanical}
M.~H{\"u}lsmann, O.~A. Peters, and P.~M. Dummer, ``Mechanical preparation of root canals: shaping goals, techniques and means,'' \emph{Endodontic topics}, vol.~10, no.~1, pp. 30--76, 2005.

\bibitem{gutmann2010problem}
J.~L. Gutmann and P.~E. Lovdahl, \emph{Problem solving in endodontics: prevention, identification and management}.\hskip 1em plus 0.5em minus 0.4em\relax Elsevier Health Sciences, 2010.

\bibitem{jafarzadeh2007ledge}
H.~Jafarzadeh and P.~V. Abbott, ``Ledge formation: review of a great challenge in endodontics,'' \emph{Journal of endodontics}, vol.~33, no.~10, pp. 1155--1162, 2007.

\bibitem{liang2022evolution}
Y.~Liang and L.~Yue, ``Evolution and development: engine-driven endodontic rotary nickel-titanium instruments,'' \emph{International Journal of Oral Science}, vol.~14, no.~1, p.~12, 2022.

\bibitem{sattapan2000defects}
B.~Sattapan, G.~J. Nervo, J.~E. Palamara, and H.~H. Messer, ``Defects in rotary nickel-titanium files after clinical use,'' \emph{Journal of endodontics}, vol.~26, no.~3, pp. 161--165, 2000.

\bibitem{qiao1993robotic}
H.~Qiao, B.~Dalay, and R.~Parkin, ``Robotic peg-hole insertion operations using a six-component force sensor,'' \emph{Proceedings of the Institution of Mechanical Engineers, Part C: Journal of Mechanical Engineering Science}, vol. 207, no.~5, pp. 289--306, 1993.

\bibitem{wang2019robotic}
S.~Wang, G.~Chen, H.~Xu, and Z.~Wang, ``A robotic peg-in-hole assembly strategy based on variable compliance center,'' \emph{IEEE Access}, vol.~7, pp. 167\,534--167\,546, 2019.

\bibitem{lim1985validity}
K.~Lim and J.~Webber, ``The validity of simulated root canals for the investigation of the prepared root canal shape,'' \emph{International endodontic journal}, vol.~18, no.~4, pp. 240--246, 1985.

\bibitem{toosi2014virtual}
A.~Toosi, M.~Arbabtafti, and B.~Richardson, ``Virtual reality haptic simulation of root canal therapy,'' in \emph{Applied Mechanics and Materials}, vol. 666.\hskip 1em plus 0.5em minus 0.4em\relax Trans Tech Publ, 2014, pp. 388--392.

\bibitem{cohen1998pathways}
S.~Cohen, R.~C. Burns, and K.~Keiser, \emph{Pathways of the pulp}.\hskip 1em plus 0.5em minus 0.4em\relax Mosby St Louis, 1998, vol.~9.

\bibitem{adams2014access}
N.~Adams and P.~Tomson, ``Access cavity preparation,'' \emph{British dental journal}, vol. 216, no.~6, pp. 333--339, 2014.

\bibitem{toothsize}
V.~Paredes, J.~Gandia~Franco, and R.~Cibrian, ``Determination of bolton tooth-size ratios by digitization, and comparison with the traditional method,'' \emph{European journal of orthodontics}, vol.~28, pp. 120--5, 05 2006.

\bibitem{rootcanal}
R.~Lähdesmäki and L.~Alvesalo, ``Root lengths in 47, xyy males' permanent teeth,'' \emph{Journal of dental research}, vol.~83, pp. 771--5, 11 2004.

\bibitem{zehnder2016guided}
M.~Zehnder, T.~Connert, R.~Weiger, G.~Krastl, and S.~K{\"u}hl, ``Guided endodontics: accuracy of a novel method for guided access cavity preparation and root canal location,'' \emph{International endodontic journal}, vol.~49, no.~10, pp. 966--972, 2016.

\bibitem{lee2020three}
J.~Lee, S.-H. Lee, J.-R. Hong, K.-Y. Kum, S.~Oh, A.~S. Al-Ghamdi, F.~A. Al-Ghamdi, A.~O. Mandorah, J.-H. Jang, and S.~W. Chang, ``Three-dimensional analysis of root anatomy and root canal curvature in mandibular incisors using micro-computed tomography with novel software,'' \emph{Applied Sciences}, vol.~10, no.~12, p. 4385, 2020.

\bibitem{haapasalo2013evolution}
M.~Haapasalo and Y.~Shen, ``Evolution of nickel--titanium instruments: from past to future,'' \emph{Endodontic topics}, vol.~29, no.~1, pp. 3--17, 2013.

\bibitem{kwak2021effects}
S.~W. Kwak, J.-H. Ha, Y.~Shen, M.~Haapasalo, and H.-C. Kim, ``Effects of root canal curvature and mechanical properties of nickel-titanium files on torque generation,'' \emph{Journal of endodontics}, vol.~47, no.~9, pp. 1501--1506, 2021.

\bibitem{pettiette2001evaluation}
M.~T. Pettiette, E.~O. Delano, and M.~Trope, ``Evaluation of success rate of endodontic treatment performed by students with stainless-steel k--files and nickel--titanium hand files,'' \emph{Journal of endodontics}, vol.~27, no.~2, pp. 124--127, 2001.

\bibitem{martins2020mechanical}
J.~N. Martins, E.~J. Silva, D.~Marques, M.~R. Pereira, A.~Ginjeira, R.~J. Silva, F.~M.~B. Fernandes, and M.~A. Versiani, ``Mechanical performance and metallurgical features of protaper universal and 6 replicalike systems,'' \emph{Journal of Endodontics}, vol.~46, no.~12, pp. 1884--1893, 2020.

\bibitem{hallenberg2007robot}
J.~Hallenberg, ``Robot tool center point calibration using computer vision,'' Master's thesis, Link{\"o}ping University, Computer Vision, The Institute of Technology, 2007.

\bibitem{yang2017four}
C.~Yang, J.~Wang, L.~Mi, X.~Liu, Y.~Xia, Y.~Li, S.~Ma, and Q.~Teng, ``A four-point measurement model for evaluating the pose of industrial robot and its influence factor analysis,'' \emph{Industrial Robot: An International Journal}, vol.~44, no.~3, pp. 343--352, 2017.

\bibitem{fu1983robotics}
K.~S. Fu, R.~C. Gonzalez, and C.~G. Lee, \emph{Robotics}.\hskip 1em plus 0.5em minus 0.4em\relax IEEE Computer Society Press, 1983.

\bibitem{lewis1993control}
F.~Lewis, C.~Abdallah, and D.~Dawson, ``Control of robot,'' \emph{Manipulators, Editorial Maxwell McMillan, Canada}, pp. 25--36, 1993.

\bibitem{vougioukas2001bias}
S.~Vougioukas, ``Bias estimation and gravity compensation for force-torque sensors,'' in \emph{Proceedings of 3rd WSEAS Symposium on Mathematical Methods and Computational Techniques in Electrical Engineering. Athens, Greece: WSEAS Press}.\hskip 1em plus 0.5em minus 0.4em\relax Citeseer, 2001, pp. 82--85.

\bibitem{geng19943}
Z.~J. Geng and L.~S. Haynes, ``A “3-2-1” kinematic configuration of a stewart platform and its application to six degree of freedom pose measurements,'' \emph{Robotics and computer-integrated manufacturing}, vol.~11, no.~1, pp. 23--34, 1994.

\bibitem{thomas2005performance}
F.~Thomas, E.~Ottaviano, L.~Ros, and M.~Ceccarelli, ``Performance analysis of a 3-2-1 pose estimation device,'' \emph{IEEE Transactions on Robotics}, vol.~21, no.~3, pp. 288--297, 2005.

\bibitem{ottaviano2010application}
E.~Ottaviano, M.~Ceccarelli, and F.~Palmucci, ``An application of catrasys, a cable-based parallel measuring system for an experimental characterization of human walking,'' \emph{Robotica}, vol.~28, no.~1, pp. 119--133, 2010.

\bibitem{jeong1999kinematics}
J.~W. Jeong, S.~H. Kim, and Y.~K. Kwak, ``Kinematics and workspace analysis of a parallel wire mechanism for measuring a robot pose,'' \emph{Mechanism and Machine Theory}, vol.~34, no.~6, pp. 825--841, 1999.

\bibitem{raghavan1993stewart}
M.~Raghavan, ``The {S}tewart platform of general geometry has 40 configurations,'' \emph{Journal of Mechanical Design}, vol. 115, no.~2, pp. 277--282, 1993.

\bibitem{nguyen1991efficient}
C.~C. Nguyen, Z.-L. Zhou, S.~S. Antrazi, and C.~Campbell, ``Efficient computation of forward kinematics and jacobian matrix of a stewart platform-based manipulator,'' in \emph{IEEE Proceedings of the SOUTHEASTCON'91}.\hskip 1em plus 0.5em minus 0.4em\relax IEEE, 1991, pp. 869--874.

\bibitem{safeena2022survey}
M.~Safeena, K.~Jiji \emph{et~al.}, ``Survey paper on the forward kinematics solution of a stewart platform,'' in \emph{2022 Second International Conference on Next Generation Intelligent Systems (ICNGIS)}.\hskip 1em plus 0.5em minus 0.4em\relax IEEE, 2022, pp. 1--8.

\bibitem{gibson2016principles}
R.~F. Gibson, \emph{Principles of composite material mechanics}.\hskip 1em plus 0.5em minus 0.4em\relax CRC press, 2016.

\bibitem{winkler2015implicit}
A.~Winkler and J.~Such{\`y}, ``Implicit force control of a position controlled robot--a comparison with explicit algorithms,'' \emph{International Journal of Computer and Information Engineering}, vol.~9, no.~6, pp. 1447--1453, 2015.

\bibitem{augugliaro2013admittance}
F.~Augugliaro and R.~D'Andrea, ``Admittance control for physical human-quadrocopter interaction,'' in \emph{2013 European Control Conference (ECC)}.\hskip 1em plus 0.5em minus 0.4em\relax IEEE, 2013, pp. 1805--1810.

\bibitem{350927}
H.~Seraji, ``Adaptive admittance control: an approach to explicit force control in compliant motion,'' in \emph{Proceedings of the 1994 IEEE International Conference on Robotics and Automation}, 1994, pp. 2705--2712 vol.4.

\bibitem{canny1983finding}
J.~F. Canny, ``Finding edges and lines in images,'' Master's thesis, Massachusetts Institute of Technology, 1983.

\bibitem{phasespace}
A.~Aristidou and J.~Lasenby, ``Motion capture with constrained inverse kinematics for real-time hand tracking,'' in \emph{2010 4th International Symposium on Communications, Control and Signal Processing (ISCCSP)}.\hskip 1em plus 0.5em minus 0.4em\relax IEEE, 2010, pp. 1--5.

\end{thebibliography}

\end{document}